\documentclass[11pt]{article}
\usepackage[final]{acl}
\usepackage{times}
\usepackage{latexsym}
\usepackage[T1]{fontenc}
\usepackage[utf8]{inputenc}
\usepackage{microtype}
\usepackage{inconsolata}
\usepackage{booktabs}
\usepackage{multirow}
\usepackage{amsmath}
\usepackage{graphicx}
\usepackage{dirtytalk}
\usepackage{pifont}
\usepackage{enumitem}
\usepackage{adjustbox}
\usepackage{xcolor}
\usepackage{url}
\usepackage{makecell}

\title{Dialects of Translationese Shape Language Model Learning}

\author{
  Jenny Kunz
  \\
  Department of Computer and Information Science
  \\
  Linköping University
  \\
  \texttt{jenny.kunz@liu.se}
}

\begin{document}
\maketitle
\begin{abstract}

Machine-translated data is widely used in multilingual NLP, particularly where native text is scarce. However, translated text differs systematically from native text. This phenomenon is known as \textit{translationese}, and it reflects both traces of the source language and characteristic properties of translation itself. 
In this paper, we study how training on machine-translated data affects small English language models, focusing on how translationese from different source languages shapes linguistic acceptability judgments and language modeling for different domains. We train models on English text translated from 24 typologically and resource-diverse source languages, enabling a systematic analysis of how source language and corpus properties influence what models learn. 
Our results show that the source language has a clear impact on model behavior: general perplexity is more driven by the lexical diversity of the translated corpus, but grammatical performance is strongly correlated to typological similarity to English if trained on enough data. Even translation quality is a strong predictor of language modeling performance.\footnote{The models are available at: \url{https://huggingface.co/collections/jekunz/translationese-english-models}}

\end{abstract}

\section{Introduction}

Translated text is everywhere in multilingual NLP, particularly for low-resource languages. Studies have shown that a high share of web-crawled data is machine-translated, especially for lower-resourced languages \citep{thompson-etal-2024-shocking}. Translated data is also commonly used in instruction tuning \citep{muennighoff-etal-2023-crosslingual, chen-etal-2024-monolingual}. 

However, translated text differs systematically from native text, even when done by professional human translators. These differences, known as \textit{translationese}, include traces of the source language and characteristic stylistic and structural patterns \citep{gellerstam86, baker}. Even human translations resemble non-native language more than native production \citep{rabinovich-etal-2016-similarities}, and machine translation is known to reduce lexical and morphological richness \citep{vanmassenhove-etal-2021-machine, niu2024}. 

Translated data is nonetheless very useful, and including it in pre-training can improve model performance for lesser-resourced languages \citep{sennrich-etal-2016-improving, wang-etal-2025-multilingual-language}. \citet{doshi-etal-2024-pretraining} show that adding data translated from English helps for Indic languages. However, in control experiments in the other translation direction, they also find that training on text translated to English degrades linguistic acceptability compared to native English, and that the source language of the translation can make a difference. We build on these findings, studying how translationese from 24 diverse source languages shapes the linguistic knowledge of small language models. We evaluate language modeling performance on different domains and intrinsic linguistic acceptability preferences for different syntactic and morphological phenomena. 

To this end, we train models on English text translated from 24 typologically diverse languages using the FineTranslations dataset \citep{penedo2026finetranslations}. This setup reverses the typical translation direction (English $\rightarrow$ other language) but enables evaluation on consistent, high-quality English benchmarks. It provides an upper bound for the performance of models trained on translated data, as machine translations into English are easier than into smaller languages due to the models' higher proficiency in English. Following the Goldfish setup \citep{chang2024goldfishmonolinguallanguagemodels}, we train small models with varying data scale. 
Our findings are: 
\begin{enumerate}[label=\textbullet,leftmargin=1em, itemsep=0pt, topsep=2pt]
    \item Translationese has a flattening effect on the resulting models, increasing perplexity on native text and reducing linguistic acceptability scores. 
    \item Surface lexical diversity in the translated corpus correlates with lower perplexity in low-data settings, but is not strongly correlated with linguistic acceptability performance.
    \item Typological similarity between the source language and English significantly improves grammatical performance given enough data. 
    \item Models trained on typologically similar source languages show lower mutual perplexity on each other's pretraining data, indicating that their \say{dialects} \citep{koppel-ordan-2011-translationese} of translationese are also similar. This shows that translationese English is not just noisy English, but has structured variation that depends on the source.
    \item Translation quality scores of the model used for the translations are highly predictive of language modeling but not of grammatical performance.
\end{enumerate}

\section{Related Work}
\label{sec:rel_work}

The term \textit{translationese} was introduced by \citet{gellerstam86} to describe systematic properties of translated text. 
Translators follow certain norms, behaviors, and policies that shape the text \citep{EvenZohar1978Position}. 
(Human-) Translated texts tend to be lexically and syntactically simpler, more explicit, and more conventional in style \citep{baker}. 
Translations also carry traces of the source language, similar to cross-linguistic influence in multilingual speakers \citep{jarvis2008crosslinguistic}.

Although first studied in human translation, translationese also arises in machine translation (MT). Especially earlier MT systems, from statistical systems like Moses \citep{koehn-etal-2007-moses} to specialized neural systems like OPUS-MT \citep{tiedemann-thottingal-2020-opus},
often produce literal translations that closely followed source-language structure. Machine-translated text has been shown to be less structurally diverse than human translations \citep{luo-etal-2024-diverge, niu2024} and to lose lexical and morphological richness compared to native text \citep{vanmassenhove-etal-2021-machine}.

Translationese is easy to detect automatically. Simple classifiers can reliably distinguish translated from native text, and even reliably determine the source language of a translated text, which lead \citet{koppel-ordan-2011-translationese} to introducing the notion of \say{dialects} of translationese, which we adopt in this work. 
Even unsupervised clustering separates them within the same domain \citep{rabinovich-wintner-2015-unsupervised}. Interestingly, classifiers outperform professional translators at this task, with function words, morphosyntactic categories, and pronouns being particularly informative features \citep{baroni-bernadini-2005}.

Several approaches aim to reduce translationese in MT output, including models trained to produce more natural target-language text \citep{riley-etal-2020-translationese}. Large language models have also been shown to produce more fluent, less literal translations than previous MT systems \citep{raunak-etal-2023-gpts}. Nevertheless, even high-quality MT output differs detectably from native text, and there is a trade-off between fluency and accuracy, with accuracy typically prioritized \citep{lim-etal-2024-simpsons}. 

The use of translated data in NLP raises concerns for both evaluation and training. Translated test sets tend to be easier than native ones, which inflates performance estimates on multilingual benchmarks \citep{chen-etal-2024-good-data}. 
The pre-training data of all state-of-the-art LLMs likely contains a large share of machine-translated data, as a lot of web-crawled data is machine-translated, especially for lower-resourced languages \citep{thompson-etal-2024-shocking}, and this share is expected to increase quickly given the current technical development. Translated data has also been used explicitly to enhance pre-training corpora with synthetic data \citep{doshi-etal-2024-pretraining, wang-etal-2025-multilingual-language, kiyomaru-etal-2026-scaling, kraus2026klettermixclimbinghighqualitygerman}. 
It is also widely used particularly in instruction tuning, since few languages have large native corpora \citep{holmstrom-doostmohammadi-2023-making, muennighoff-etal-2023-crosslingual, chen-etal-2024-monolingual}. However, models instruction-tuned on translated data develop weaker preferences for natural and idiomatic language \citep{kunz2026preferencesidiomaticlanguageacquired, kunz-etal-2026-dataset}. 

Building on this literature, we systematically study how translationese from different source languages affects the syntactic and morphological knowledge acquired by small language models.



\section{Experimental Setup} 

\subsection{Models}

To train our models, we follow the Goldfish setup \citep{chang2024goldfishmonolinguallanguagemodels}, which is a suite of small monolingual Transformer models trained on controlled data budgets. All models used in this paper have 125M parameters and a 512-token context window. 
To ensure comparability of perplexity scores, we use the same tokenizer, taken from the original English Goldfish model, for all experiments. It is monolingual with a vocabulary of 50,000 tokens. The code is available at  \url{https://github.com/jekunz/dialects-translationese}.

\subsection{Languages}

\begin{figure}
    \centering
    \includegraphics[width=\linewidth]{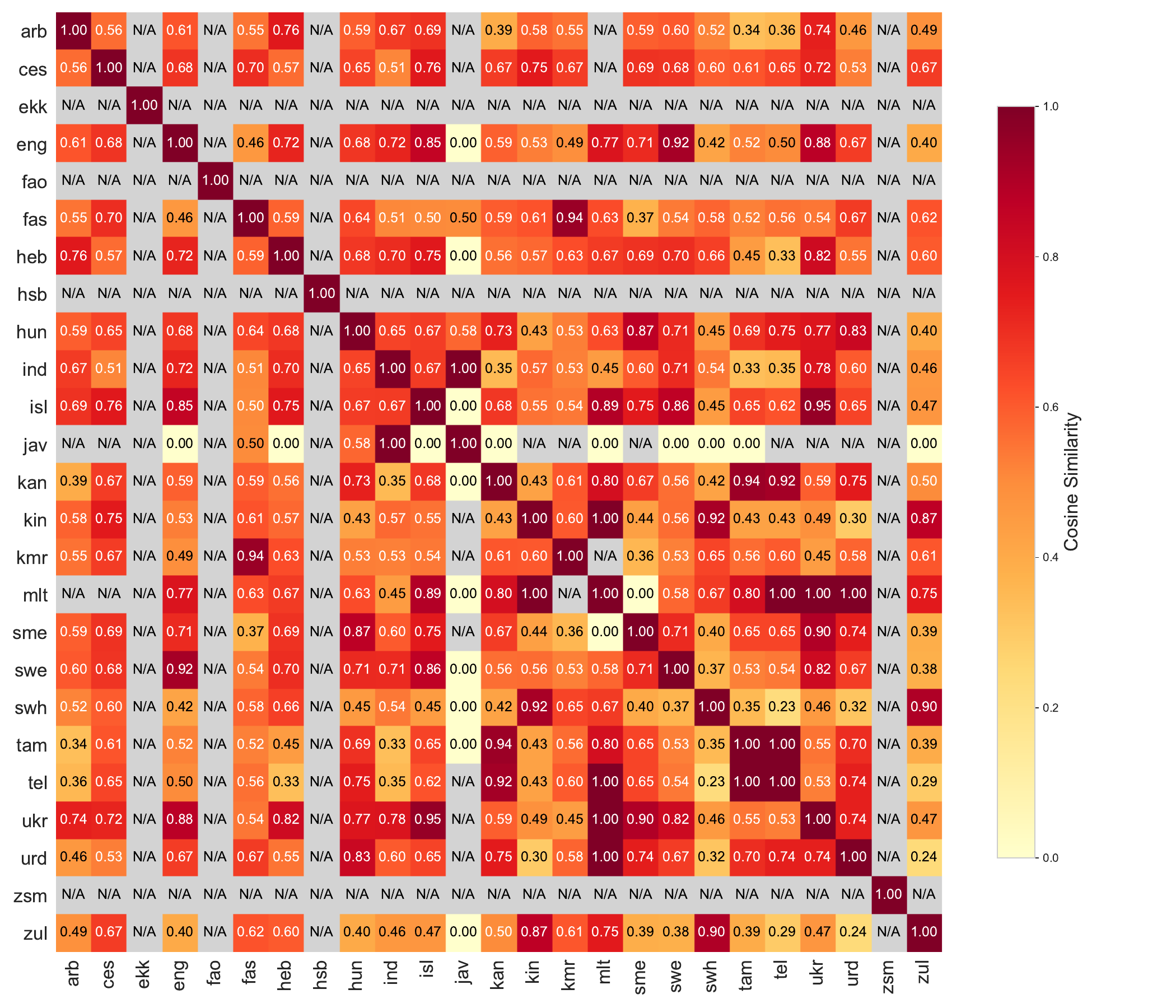}
    \caption{WALS \citep{wals} syntax feature cosine similarity, generated with lang2vec \citep{littell-etal-2017-uriel} for all language pairs. Some languages are unavailable or do not have common features (N/A).}
    \label{fig:lang2vec_wals_syntax}
\end{figure}

We include 24 typologically and genealogically diverse languages. From the Indo-European language family, we include Germanic languages (Swedish, Icelandic, Faroese), Slavic languages (Ukrainian, Czech, Upper Sorbian) and Indo-Iranian languages (Persian, Urdu, Northern Kurdish). We also include languages from the Uralic (Hungarian, Estonian, Northern Sámi), Semitic (Standard Arabic, Hebrew, Maltese), Dravidian (Tamil, Telugu, Kannada), Austronesian (Indonesian, Standard Malay, Javanese), and Niger-Congo/Bantu (Swahili, Kinyarwanda, Zulu) families. 
Within each group, we deliberately include both higher- and lower-resource languages, but with a minimum of about 100MB of available data. 
To analyze syntactic similarity to English and to each other, we compute cosine similarity over WALS syntactic features using lang2vec \citep{littell-etal-2017-uriel}. This allows us to relate model behavior to typological distance for the covered languages. Results are in Figure~\ref{fig:lang2vec_wals_syntax}. 

\subsection{Data}

We train on parts of the FineTranslations dataset \citep{penedo2026finetranslations}, which provides machine translations of FineWeb-2 \citep{penedo2025fineweb2pipelinescale} into English.
We adopt the Goldfish paper setups that train on 100MB and 1GB of data.\footnote{While in this paper we only use the 100MB and 1GB models, we also train and release the smaller variants 5MB and 10MB for completeness and future research.}
As native English baselines, we train English models on portions of FineWeb \citep{penedo2024the}. 

\paragraph{Statistics}

In Table~\ref{tab:merged_dataset_stats}, we analyze the translated 100MB corpora using surface-level statistics: we report the size of the base corpus in number of words and disk space; even though the training corpora used in our experiments are controlled to have the same size, they are sampled from source corpora whose sizes differ across languages. This can lead to differences in diversity and coverage, which could affect the resulting models.
We also analyze the vocabulary size (whitespace-tokenized), type–token ratio (TTR), bigram TTR, and average sentence length of the FineTranslations portion that we use for training.\footnote{In Appendix~\ref{app:morph_complexity}, we also analyze experimental measures of morphological complexity of source and target corpora.} 
The English FineWeb baseline shows the highest diversity, measured in vocabulary size, TTR and bigram TTR. Among English-translated corpora, vocabulary size ranges from 350k to 610k types. Corpora translated from Hungarian and Javanese exhibit relatively high TTR and bigram TTR, while the translation from Kinyarwanda shows the lowest diversity in all metrics. 
The average sentence length also varies substantially (8.8--20.9 tokens). Translations from smaller languages such as Northern Sámi, Faroese, and Upper Sorbian have shorter average sentence lengths. Translations from Arabic and Kinyarwanda show the longest sentences, followed by translations from Persian, Urdu, and Hebrew.

We further explore how predictive base corpus size is for diversity in the 100MB translated corpus. 
For the 100MB datasets, correlations between base corpus size and TTR are weak and not statistically significant\footnote{All reported p-values in this paper are two-tailed. Statistical significance was assessed at $\alpha = 0.05$.} (Pearson $r = 0.238$, $p=0.2636$; Spearman $\rho = +0.177$, $p=0.4070$). 
Bigram TTR showed a moderate positive correlation (Pearson $r = 0.410$, $p = 0.0468$; Spearman $\rho = +0.445$, $p=0.02925$). 
For the 1GB datasets, correlations are stronger for TTR (Pearson $r=0.439$, $p=0.0603$; Spearman $\rho = 0.439$, $p = 0.0603$) and significant for bigram TTR (Pearson $r=0.579$, $p=0.0094$; Spearman $\rho = 0.681$, $p = 0.0013$). This suggests that in larger base corpora, the diversity of translated text increases.

\begin{table*}[t]
\centering
\footnotesize 
\begin{tabular}{llrrrrrr}
\toprule
\multirow{2}{*}{Source} & \multirow{2}{*}{Language}
& \multicolumn{2}{c}{Base Corpus}
& \multicolumn{4}{c}{Translated Corpus} \\
\cmidrule(lr){3-4} \cmidrule(lr){5-8}
& & Words & UTF-8 Bytes
& VocabSize & TTR (\%) & BigrTTR (\%) & AvgSenLen \\
\midrule
eng & English & 18{,}500{,}000M & -- & 835517 & 4.77 & 35.42 & 13.69 \\ \midrule
arb & Standard Arabic & 32{,}813M & 293.59\,GB & 516177 & 3.02 & 25.88 & 20.87 \\
ces & Czech & 35{,}479M & 206.33\,GB & 590228 & 3.42 & 29.19 & 13.63 \\
ekk & Estonian & 6{,}564M & 40.82\,GB & 553751 & 3.25 & 28.36 & 13.32 \\
fao & Faroese & 101M & 564.10\,MB & 448266 & 2.55 & 21.25 & 10.76 \\
fas & Persian & 39{,}706M & 304.62\,GB & 435917 & 2.56 & 24.11 & 18.30 \\
heb & Hebrew & 8{,}463M & 68.71\,GB & 470477 & 2.68 & 24.85 & 17.09 \\
hsb & Upper Sorbian & 15M & 92.13\,MB & 397972 & 2.60 & 19.76 & 11.89 \\
hun & Hungarian & 30{,}919M & 199.69\,GB & 561913 & 3.33 & 29.63 & 14.42 \\
ind & Indonesian & 60{,}264M & 348.65\,GB & 550569 & 3.22 & 28.53 & 14.53 \\
isl & Icelandic & 1{,}696M & 10.27\,GB & 481243 & 2.77 & 24.95 & 13.84 \\
jav & Javanese & 140M & 783.32\,MB & 611377 & 3.58 & 28.93 & 12.33 \\
kan & Kannada & 748M & 12.91\,GB & 494342 & 2.90 & 25.94 & 13.47 \\
kin & Kinyarwanda & 128M & 793.30\,MB & 350870 & 1.98 & 18.69 & 20.89 \\
kmr & Northern Kurdish & 221M & 1.16\,GB & 495600 & 2.85 & 23.83 & 15.57 \\
mlt & Maltese & 287M & 1.53\,GB & 480824 & 2.76 & 24.28 & 14.24 \\
sme & Northern Sámi & 25M & 177.92\,MB & 419365 & 2.52 & 20.30 & 8.75 \\
swe & Swedish & 35{,}745M & 202.96\,GB & 509665 & 2.88 & 26.75 & 14.27 \\
swh & Swahili & 570M & 3.08\,GB & 492842 & 2.83 & 24.23 & 16.32 \\
tam & Tamil & 1{,}937M & 36.97\,GB & 532375 & 3.11 & 26.29 & 12.74 \\
tel & Telugu & 891M & 14.42\,GB & 552762 & 3.21 & 26.96 & 12.09 \\
ukr & Ukrainian & 25{,}586M & 254.86\,GB & 594748 & 3.54 & 27.25 & 13.67 \\
urd & Urdu & 2{,}733M & 19.93\,GB & 417644 & 2.43 & 23.81 & 18.21 \\
zsm & Standard Malay & 5{,}648M & 31.94\,GB & 572369 & 3.28 & 27.87 & 13.20 \\
zul & Zulu & 71.7M & 487.38\,MB & 526206 & 3.00 & 27.10 & 14.83 \\
\bottomrule
\end{tabular}
\caption{Base corpus sizes (FineWeb-2) and English-translated corpus (FineTranslations) surface statistics. }
\label{tab:merged_dataset_stats}
\end{table*}

\paragraph{Translation quality}
We also analyze the relationship between translation quality and model performance. FineTranslations was translated using Gemma-3-27B \citep{gemma_2025}. We evaluate it using the massively multilingual parallel dataset BOUQuET \citep{theomnilingualmtteam2025bouquetdatasetbenchmarkopen}, from each source language into English. 
We report both the character-based chrF metric \citep{popovic-2015-chrf} and the word-based BLEU score \citep{papineni-etal-2002-bleu}, as, while the former is considered the better translation metric, BLEU offers complementary insights \citep{kumar2026evaluatingextremelylowresourcemachine}.
In addition, we measure Gemma's competence in each source language using perplexity, also computed on BOUQuET. Full scores are in Table~\ref{tab:translation_quality} in Appendix~\ref{app:gemma_quality}.

\subsection{Evaluation}

\begin{table*}[t]
\centering
\footnotesize
\setlength{\tabcolsep}{4pt}
\begin{tabular}{l cccc cccc cc}
\toprule
& \multicolumn{8}{c}{Perplexity ($\downarrow$)} & \multicolumn{2}{c}{BLiMP ($\uparrow$)} \\
\cmidrule(lr){2-9} \cmidrule(lr){10-11}
& \multicolumn{4}{c}{100MB} & \multicolumn{4}{c}{1GB} & 100MB & 1GB \\
\cmidrule(lr){2-5} \cmidrule(lr){6-9} \cmidrule(lr){10-11}
Source
& Books & Voyage & News & FineWeb
& Books & Voyage & News & FineWeb
& \multicolumn{2}{c}{Overall}\\
\midrule
eng & 19.05 & 24.10 & 18.41 & 29.06 & 12.69 & 15.62 & 11.75 & 15.25 & 67.14 & 75.44 \\
\midrule
arb & 21.94 & 30.47 & 19.79 & 46.47 & 14.94 & 19.58 & 13.53 & 23.53 & 65.14 & 70.89 \\
ces & 20.09 & 26.05 & 19.85 & 40.13 & 14.39 & 17.62 & 13.71 & 22.26 & 66.08 & 73.33 \\
ekk & 20.47 & 26.98 & 20.56 & 41.98 & 14.24 & 17.69 & 14.25 & 22.33 & 63.98 & 72.31 \\
fao & 23.79 & 30.11 & 23.99 & 51.56 & \textcolor{lightgray}{19.46} & \textcolor{lightgray}{24.68} & \textcolor{lightgray}{19.44} & \textcolor{lightgray}{34.73} & 65.66 & \textcolor{lightgray}{70.61} \\
fas & 21.71 & 31.76 & 21.42 & 48.51 & 14.97 & 20.99 & 14.42 & 24.84 & 63.38 & 71.26 \\
heb & 21.95 & 29.61 & 22.29 & 46.80 & 14.49 & 18.90 & 14.61 & 22.67 & 65.08 & 73.11 \\
hsb & 27.88 & 36.37 & 28.93 & 79.37 & \textcolor{lightgray}{26.89} & \textcolor{lightgray}{34.24} & \textcolor{lightgray}{27.59} & \textcolor{lightgray}{73.30} & 63.29 & \textcolor{lightgray}{62.82} \\
hun & 20.70 & 27.48 & 19.86 & 44.63 & 14.41 & 18.18 & 13.66 & 23.06 & 65.11 & 72.90 \\
ind & 21.15 & 28.36 & 20.44 & 39.40 & 14.46 & 18.79 & 13.63 & 21.61 & 66.39 & 72.48 \\
isl & 20.81 & 26.63 & 20.38 & 40.02 & 14.67 & 17.96 & 14.13 & 22.01 & 65.68 & 73.37 \\
jav & 21.25 & 28.18 & 23.00 & 46.95 & 15.53 & 19.08 & 15.40 & 24.68 & 65.30 & 68.95 \\
kan & 23.10 & 32.52 & 21.59 & 48.96 & 16.05 & 21.49 & 14.65 & 25.89 & 66.83 & 72.28 \\
kin & 28.21 & 39.59 & 23.81 & 63.50 & \textcolor{lightgray}{20.25} & \textcolor{lightgray}{27.97} & \textcolor{lightgray}{17.60} & \textcolor{lightgray}{34.86} & 65.18 & \textcolor{lightgray}{71.64} \\
kmr & 21.15 & 31.07 & 21.37 & 47.53 & 15.08 & 19.98 & 14.87 & 25.14 & 66.24 & 72.68 \\
mlt & 20.66 & 25.91 & 19.99 & 39.32 & 14.37 & 16.87 & 13.44 & 21.34 & 65.06 & 71.84 \\
sme & 27.42 & 33.78 & 28.38 & 65.10 & \textcolor{lightgray}{24.56} & \textcolor{lightgray}{31.43} & \textcolor{lightgray}{25.77} & \textcolor{lightgray}{55.32} & 67.26 & \textcolor{lightgray}{66.89} \\
swe & 21.39 & 26.40 & 21.05 & 38.22 & 14.40 & 17.10 & 14.21 & 20.96 & 67.67 & 74.37 \\
swh & 21.57 & 28.38 & 19.44 & 42.84 & 15.04 & 18.71 & 13.28 & 22.62 & 66.00 & 72.77 \\
tam & 22.61 & 32.17 & 20.60 & 46.80 & 15.90 & 21.43 & 14.09 & 25.19 & 67.54 & 73.49 \\
tel & 23.24 & 32.28 & 21.21 & 46.32 & 15.86 & 21.27 & 14.65 & 24.51 & 66.94 & 73.36 \\
ukr & 22.22 & 30.46 & 22.07 & 47.93 & 15.61 & 20.65 & 15.22 & 25.95 & 64.53 & 71.48 \\
urd & 22.40 & 31.38 & 19.66 & 48.49 & 15.41 & 20.07 & 13.28 & 25.08 & 65.36 & 73.14 \\
zsm & 22.26 & 30.02 & 21.09 & 44.14 & 15.35 & 19.19 & 14.01 & 22.47 & 65.96 & 72.43 \\
zul & 20.70 & 26.37 & 20.41 & 40.87 & \textcolor{lightgray}{16.12} & \textcolor{lightgray}{20.09} & \textcolor{lightgray}{15.38} & \textcolor{lightgray}{25.67} & 66.96 & \textcolor{lightgray}{70.46} \\
\bottomrule
\end{tabular}
\caption{Perplexity on FLORES+ domains and English FineWeb (10MB eval set), and overall BLiMP accuracy (\%) for 100MB and 1GB models. Results in \textcolor{lightgray}{gray} indicate that the translated corpora for these low-resource languages were smaller than 1GB. For per-category BLiMP results, see Table~\ref{tab:blimp_by_category_1GB} in Appendix~\ref{sec:appendix_full_blimp}.}
\label{tab:perplexity_blimp_combined}
\end{table*}

We evaluate the models in three ways:
\paragraph{Perplexity on native English text}
To assess general language modelling performance across multiple domains, we compute perplexity on the English source side of FLORES+ \citep{Costa-jussa2024}, which includes \textit{Wikinews}, \textit{Wikivoyage}, and \textit{Wikibooks} domains, and on a 10MB subset of native English Fineweb text.

\paragraph{Linguistic acceptability}
To evaluate syntactic and morphological phenomena in isolation, we evaluate on the BLiMP dataset \citep{warstadt-etal-2020-blimp-benchmark}, which contains minimal pairs of correct sentences and sentences containing grammatical errors of different categories. As in the original setup, we count a sample as correct if the correct sentence is assigned a lower mean perplexity than the wrong one. For analyses of the specific lingustic phenomena that BLiMP covers, we group them into the same categories as \citet{warstadt-etal-2020-blimp-benchmark}. 

\paragraph{Cross-translation perplexity}
We measure perplexity on translated corpora originating from other source languages on the 10MB datasets as a behavioral measure of similarity of the texts translated from different translation sources. 
This metric probes structure in translationese by testing whether the source language leaves a detectable footprint in translated text that persists through translation and language model training. 

\section{Results}

\subsection{Perplexity}

General language modeling results are in Table~\ref{tab:perplexity_blimp_combined}. 

\begin{table*}[t]
\centering
\footnotesize
\begin{adjustbox}{width=\textwidth}
\begin{tabular}{llccccccccccc}
\toprule
\multirow{3}{*}{Model} & \multirow{3}{*}{Predictor} 
& \multicolumn{5}{c}{Pearson $r$} 
& \multicolumn{5}{c}{Spearman $\rho$} \\
\cmidrule(lr){3-7} \cmidrule(lr){8-12}
& & Books & Vovage & News & All & FineWeb 
& Books & Voyage & News & All & FineWeb \\
\midrule
100MB & Corpus Size 
& $-0.35$ & $-0.29$ & $-0.31$ & $-0.34$ & $-0.34$
& $-0.40$ & $-0.31$ & $-0.45^{*}$ & $-0.39$ & $-0.47^{*}$ \\
(n=24) & TTR 
& $-0.60^{**}$ & $-0.56^{**}$ & $-0.38$ & $-0.56^{**}$ & $-0.51^{*}$
& $-0.48^{*}$ & $-0.42^{*}$ & $-0.25$ & $-0.44^{*}$ & $-0.45^{*}$ \\
& lang2vec  
& $-0.02$ & $-0.17$ & $-0.07$ & $-0.11$ & $-0.17$
& $-0.11$ & $-0.29$ & $-0.03$ & $-0.22$ & $-0.34$ \\
& BLEU 
& $-0.79^{***}$ & $-0.77^{***}$ & $-0.70^{***}$ & $-0.83^{***}$ & $-0.79^{***}$
& $-0.71^{***}$ & $-0.79^{***}$ & $-0.58^{**}$ & $-0.78^{***}$ & $-0.70^{***}$ \\

& chrF2 
& $-0.81^{***}$ & $-0.74^{***}$ & $-0.74^{***}$ & $-0.84^{***}$ & $-0.80^{***}$
& $-0.69^{***}$ & $-0.78^{***}$ & $-0.57^{**}$ & $-0.77^{***}$ & $-0.68^{***}$ \\

& BOUQuET PPL
& $+0.62^{**}$ & $+0.38$ & $+0.81^{***}$ & $+0.63^{**}$ & $+0.65^{**}$
& $+0.54^{**}$ & $+0.51^{*}$ & $+0.50^{*}$ & $+0.55^{**}$ & $+0.48^{*}$ \\
\midrule
1GB & Corpus Size 
& $-0.44$ & $-0.17$ & $-0.21$ & $-0.28$ & $-0.29$
& $-0.42$ & $-0.15$ & $-0.21$ & $-0.27$ & $-0.27$ \\
(n=19) & TTR 
& $+0.07$ & $-0.07$ & $+0.33$ & $+0.06$ & $-0.01$
& $+0.04$ & $-0.09$ & $+0.22$ & $+0.02$ & $-0.01$ \\
& lang2vec  
& $-0.45$ & $-0.35$ & $-0.31$ & $-0.42$ & $-0.43$
& $-0.53^{*}$ & $-0.50^{*}$ & $-0.14$ & $-0.52^{*}$ & $-0.45$ \\
& BLEU 
& $-0.79^{***}$ & $-0.87^{***}$ & $-0.70^{***}$ & $-0.81^{***}$ & $-0.76^{***}$
& $-0.86^{***}$ & $-0.90^{***}$ & $-0.60^{**}$ & $-0.89^{***}$ & $-0.76^{***}$ \\

& chrF2 
& $-0.84^{***}$ & $-0.90^{***}$ & $-0.76^{***}$ & $-0.86^{***}$ & $-0.82^{***}$
& $-0.87^{***}$ & $-0.90^{***}$ & $-0.60^{**}$ & $-0.89^{***}$ & $-0.77^{***}$ \\

& BOUQuET PPL
& $+0.81^{***}$ & $+0.73^{***}$ & $+0.84^{***}$ & $+0.81^{***}$ & $+0.85^{***}$
& $+0.61^{**}$ & $+0.59^{**}$ & $+0.40$ & $+0.58^{**}$ & $+0.44^{*}$ \\
\bottomrule
\end{tabular}
\end{adjustbox}
\caption{Correlations between predictor variables and perplexity (FLORES+ subsets and FineWeb) across model sizes. Significance levels: $^{***}p<0.001$, $^{**}p<0.01$, $^{*}p<0.05$. 1GB excludes fao, hsb, kin, sme, zul (smaller corpora). lang2vec analyses have reduced sample sizes due to missing data (100MB $n$=20, 1GB $n$=17).}
\label{tab:ppl_correlations_combined}
\end{table*}

\subsubsection{Performance and gap to baseline}

No model trained on translated data matches the native English baseline across any domain or model size.
For the 100MB models, the gap to the English baseline is smaller on FLORES+ (1--2.5 points) but larger on FineWeb (+8.84 points). Note however that even the baseline models' training data is a part of FineWeb, so distributions of train and test data are likely more similar than for other domains. 
At 1GB, perplexity decreases substantially for all models. The gap on FLORES+ narrows to under 1 point on Wikinews, while the FineWeb gap remains large (+5.71 points). So, scaling helps overall, but does not recover native-like performance. 

\paragraph{Variation across source languages.}
Performance varies strongly across source languages. For the 100MB models, perplexity ranges from 20.09--28.21 on Wikibooks, 25.91--39.59 on Wikivoyage, 19.44--28.93 on Wikinews, and 38.22--79.37 on FineWeb. The 1GB models reduce this variance but do not eliminate it. Among FLORES+ domains, Wikivoyage shows the largest variation while Wikinews is the most stable after scaling.

\paragraph{Typological distance vs.\ performance.}
Typological similarity to English does not consistently predict lower perplexity. Typologically distant languages such as Estonian, Hungarian, Maltese, and Zulu often perform as well as or better than Germanic languages on FLORES. For instance, Estonian and Hungarian match or slightly outperform Swedish on Wikibooks at 1GB, and Maltese has the best Wikivoyage scores in both data regimes. On FineWeb, Swedish consistently has the lowest perplexity among translated models, though this may reflect domain overlap between Swedish and English web corpora rather than, or in addition to, typological proximity. 

\subsubsection{Correlation analysis} 

To better understand performance differences, we analyze correlations between perplexity and three predictors: base corpus size, lexical diversity (TTR) of the translated corpus, and syntactic similarity to English. 
Results are in Table~\ref{tab:ppl_correlations_combined}. 

For the 100MB models ($n=24$), TTR shows the largest and most consistent correlations. It correlates negatively with perplexity on both FLORES+ (Pearson $r=-0.56$ for the full dataset) and FineWeb ($r=-0.51$). 
The not significant exception is Wikinews. Corpus size shows weaker but directionally consistent negative correlations; Pearson $r$ values are not significant, but Spearman $\rho$ reaches significance on Wikinews and FineWeb, reflecting the skewed distribution of corpus sizes, where the rank-based Spearman correlations are more robust. Lang2vec similarity shows no significant correlation with perplexity at this scale.

For the 1GB models ($n=19$), the picture is different. The TTR correlation is no longer there (Pearson $r=+0.06$ on FLORES, $r=-0.01$ on FineWeb), suggesting that lexical diversity does not predict performance once sufficient training data is available\footnote{We find in an ablation (see Appendix~\ref{sec:filtering_ablation}) that the effect is reinforced, but not fully driven, by the exclusion of the low-resource languages that do not have 1GB corpora.}. Corpus size remains negatively correlated, but the effect is not significant. But most notably, syntactic similarity becomes considerably more predictive: lang2vec WALS similarity now shows moderate negative correlations on FLORES+ (average Pearson $r=-0.42$; Spearman $\rho=-0.52^{*}$, the latter significant except on Wikinews), i.e., languages syntactically closer to English are associated with lower perplexity at this scale. The effect is weaker and non-significant on FineWeb, but directionally consistent.

The analysis also shows a clear relationship between translation quality and model performance. Translation quality scores have strong negative correlations with perplexity across all test sets, with chrF2 $r$ between $-0.86$ and $-0.80$ across scales for both datasets.
This means that languages that Gemma-3-27B translates well into English produce better language models. 
Gemma's perplexity in the source language shows weaker ($r$ between $0.63$ and $0.85$) but still significant correlations, showing that model’s competence in the source language matters, although its ability to translate from the source language into the target language is the best predictor for overall language modeling quality.

\subsection{Linguistic Acceptability}
Overall BLiMP accuracy scores are in Table~\ref{tab:perplexity_blimp_combined}; per-category results are in Table~\ref{tab:blimp_by_category_1GB} in Appendix~\ref{sec:appendix_full_blimp}.

\begin{table*}[t]
\centering
\footnotesize
\setlength{\tabcolsep}{4pt}
\begin{adjustbox}{width=1.99\columnwidth}
\begin{tabular}{lcccccc}
\toprule
 & Corpus Size & TTR & lang2vec & PPL & BLEU & chrF2 \\
\midrule
100MB & $-0.09$ ($-0.09$) & $+0.11$ ($+0.08$) & $+0.06$ ($-0.10$) & $+0.34$ ($+0.40$) & $-0.37$ ($-0.37$) & $-0.40$ ($-0.39$) \\
1GB & $+0.01$ ($+0.08$) & $-0.32$ ($-0.13$) & $+0.61^{**}$ ($+0.32$) & $-0.41$ ($-0.20$) & $+0.10$ ($+0.12$) & $+0.04$ ($+0.10$) \\
\bottomrule
\end{tabular}
\end{adjustbox}
\caption{Correlations with overall BLiMP accuracy (Pearson $r$, Spearman $\rho$ in parentheses). $n$=24 for 100MB, $n$=19 for 1GB models. PPL, BLEU, and chrF2 refer to Gemma's source-language perplexity and translation scores.}
\label{tab:blimp_overall_compact}
\end{table*}

\subsubsection{Performance and gap to baseline}

For the 100MB models, the English baseline reaches 66.20\%. Translated-source models perform very similarly, ranging from 63.29\% (Upper Sorbian) to 67.67\% (Swedish); a spread of only 4.38 points, noticeably smaller than what we observed for perplexity. Swedish, the typologically closest language to English, achieves the highest translated score, just \textit{above} the baseline. However, several typologically distant languages perform nearly as well: Tamil (67.54\%), Northern Sámi (67.26\%), Zulu (66.96\%), and Kannada (66.83\%), showing that typological proximity alone does not determine performance at this scale.
In Table~\ref{tab:blimp_by_category_1GB} in Appendix~\ref{sec:appendix_full_blimp}, we see that 
performance differs across phenomena. 
Agreement categories (Anaphor Agreement, Irregular Forms, Determiner--Noun Agreement) are generally high (75--90\%), suggesting that local morphosyntactic constraints are learned reliably even from limited translated data. Island Effects are consistently the most difficult (English: 43.67\%; most translations: 40--49\%), followed by NPI Licensing and Filler--Gap dependencies (often 45--60\%), reflecting difficulty with long-distance dependencies. Some distant languages show unexpected peaks: Zulu reaches 93.10\% on Anaphor Agreement, above the English score of 91.30\%, maybe reflecting translation patterns that overrepresent certain constructions.

Scaling to 1GB improves performance across models. The baseline increases from 66.20\% to 76.01\%, and translated-source models also improve substantially but are for this scale \textit{consistently below} the baseline. Swedish again performs best (74.37\%), followed by Tamil (73.49\%), while Javanese scores lowest among fully trained models (68.95\%). 
Gains are largest for structurally complex phenomena: Island Effects improve by over 13 points for English (43.67\% $\rightarrow$ 56.76\%), with similar gains across languages, and Control/Raising and Binding improve by 6--10 points. Agreement phenomena approach ceiling performance (often above 95\%). NPI Licensing and Quantifiers remain challenging even at this scale, typically staying in the mid-50s to low-60s and mid-70s respectively. 

\subsubsection{Correlation Analysis}

We report correlations between BLiMP accuracy and corpus size, TTR, syntactic similarity to English and translation quality scores in Table~\ref{tab:blimp_overall_compact}.

For the 100MB models ($n=24$), none of the predictors corpus size, TTR, or syntactic similarity to English shows significant correlations with overall BLiMP accuracy, and effect sizes are small ($ (|r| \leq 0.11) $). This is consistent with the narrow accuracy range at this scale: in low-data settings, BLiMP scores appear insensitive to corpus properties. 
For the 1GB models ($n=19$) however, the pattern is different. While corpus size remains unrelated to BLiMP accuracy (Pearson $r=+0.01$; Spearman $\rho=+0.08$), and TTR shows a weak, non-significant negative correlation ($r=-0.32$), lang2vec similarity to English shows a strong and significant positive correlation with BLiMP accuracy ($r=+0.61$). 
This suggests that once sufficient training data is available, typological similarity to English is strongly associated with the models' grammatical competence. 

Correlations for the different BLiMP categories can be found in Table~\ref{tab:blimp_categories_combined} in Appendix~\ref{sec:appendix_full_blimp}. 
At 100MB, correlations between predictors and per-category BLiMP scores are weak and inconsistent across phenomena. At 1GB, syntactic similarity shows the largest and most consistent correlations, with particularly strong effects for agreement phenomena (Anaphor and Subject--Verb Agreement: $r=+0.74^{***}$) and long-distance dependencies (Filler--Gap: $r=+0.66^{**}$; Ellipsis: $r=+0.61^{**}$; Island Effects: $r=+0.53^{*}$). TTR correlations are mostly negative and not significant.

Correlations with the Gemma model's BOUQuET perplexity and machine translation scores are also small, as shown in Table~\ref{tab:blimp_overall_compact}. This suggests that, while translation model competence in the source language is very important for overall language modeling quality, it does not predict isolated morphological or syntactic preferences. For the latter, only typological similarity to the target language (English) is a meaningful predictor. 

\subsection{Cross-Evaluation Perplexity}

\begin{figure*}
    \centering
    \includegraphics[width=0.49\linewidth]{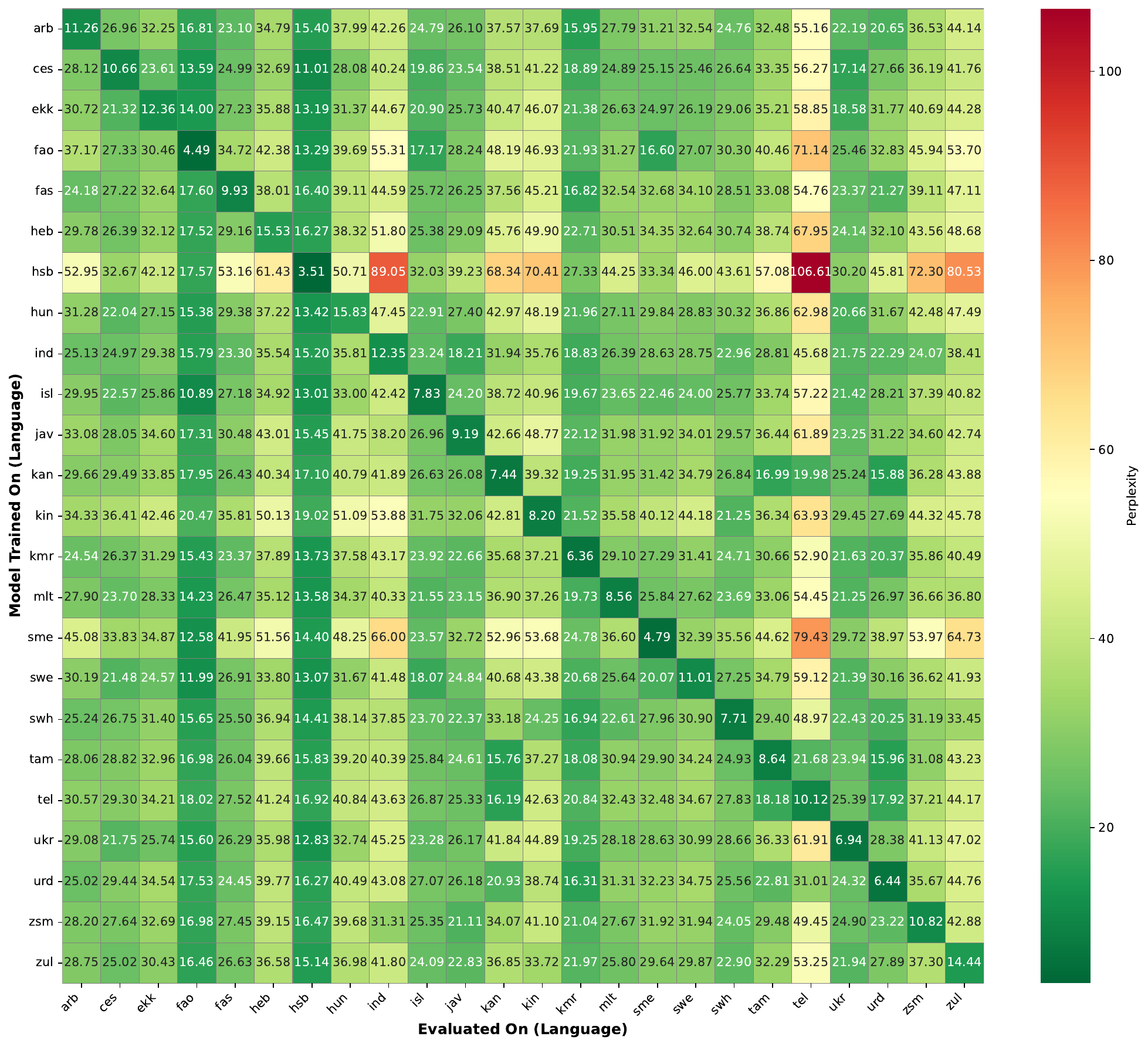}
    \includegraphics[width=0.49\linewidth]{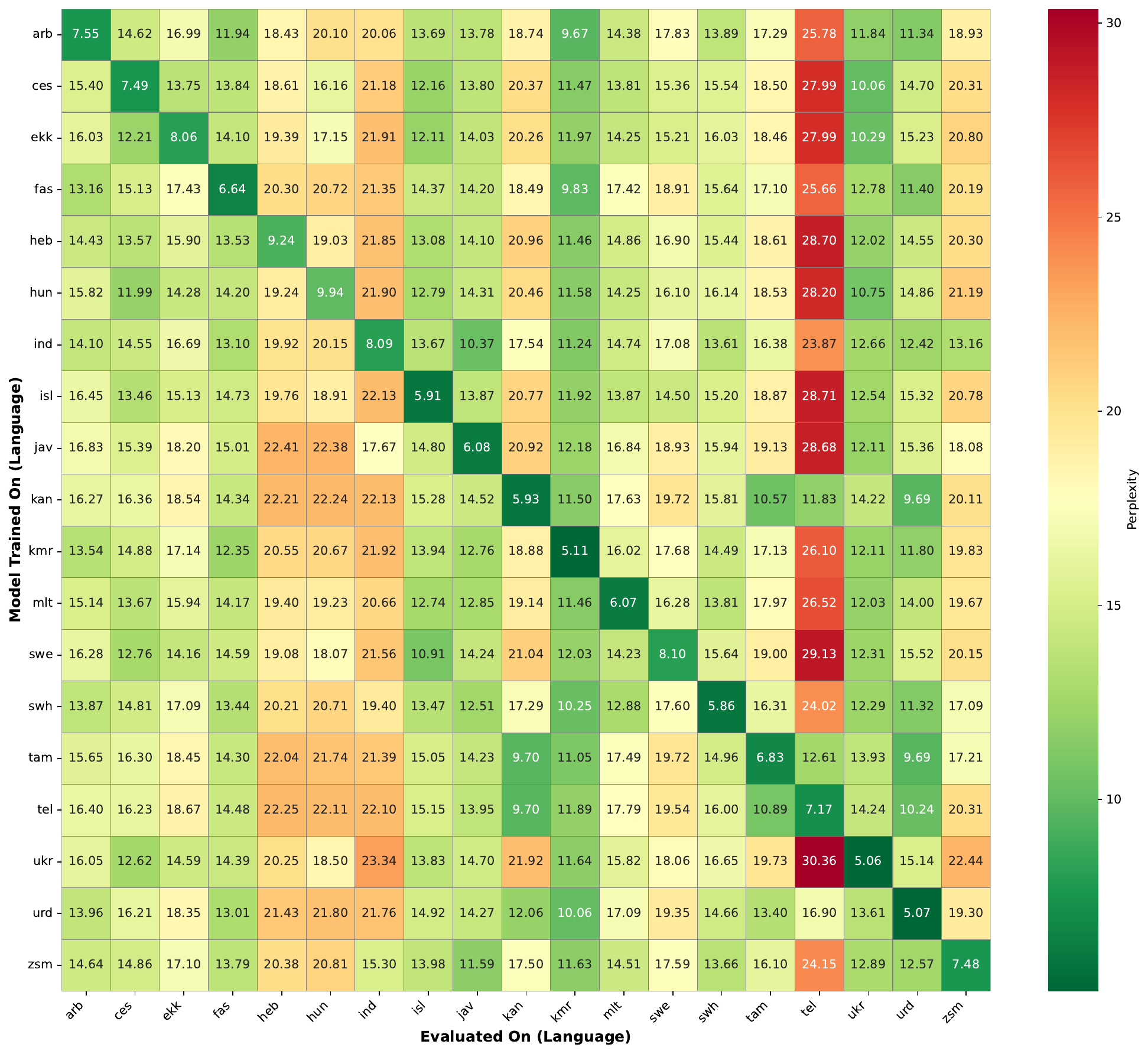}
    \caption{Cross-evaluation perplexity matrices (heatmaps) for models trained on translations from language A, evaluated on translations from language B. Left: 100MB models; right: 1GB models. }
    \label{fig:cross-eval-heatmaps}
\end{figure*}

\begin{figure}
    \centering
    \includegraphics[width=0.49\linewidth]{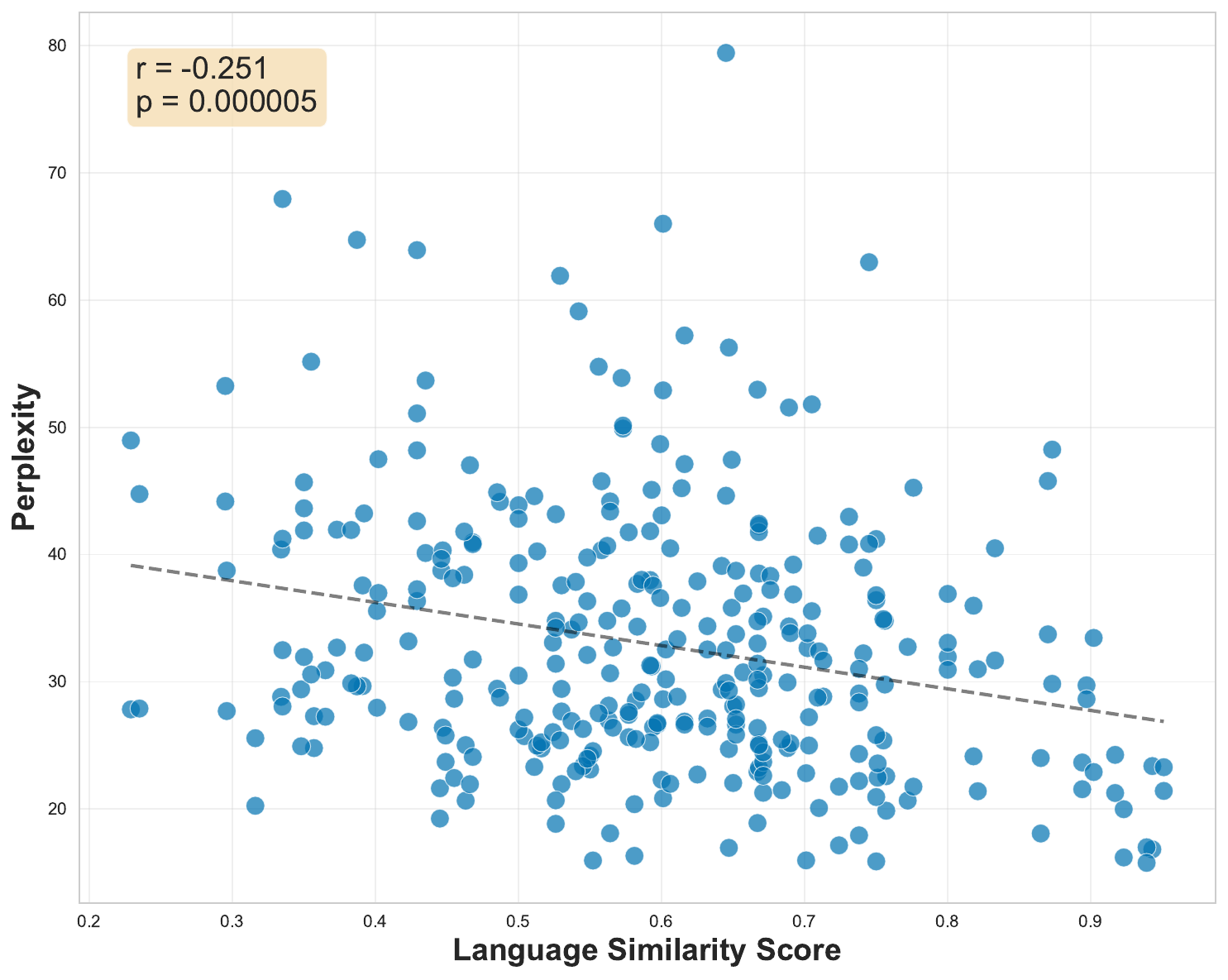}
    \includegraphics[width=0.49\linewidth]{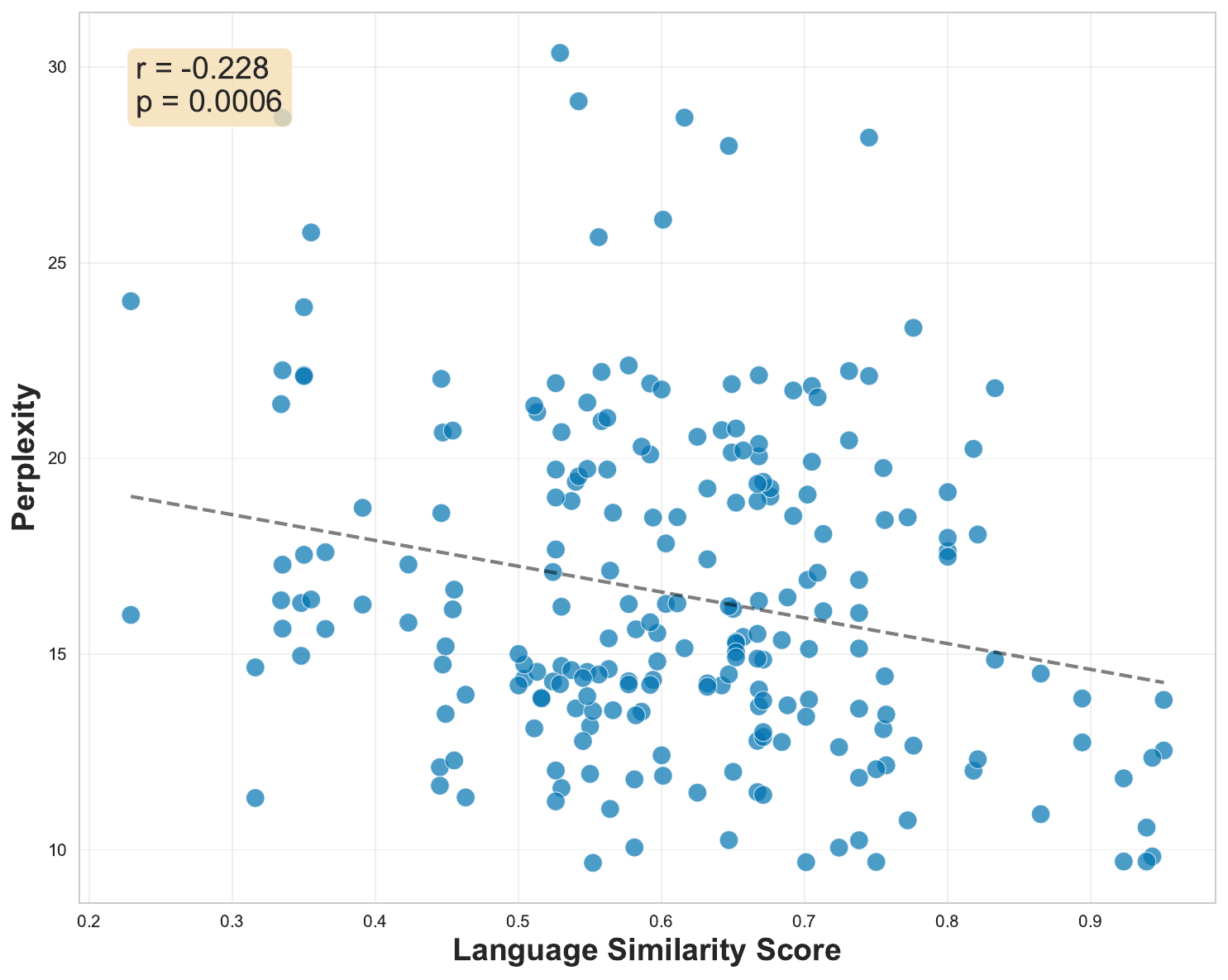}
    \caption{Correlations between language similarity and cross-evaluation perplexity across language pairs. Left: 100MB models; right: 1GB models.   
    }
    \label{fig:cross-eval-correlations}
\end{figure}

We next ask whether models trained on translations from language A generalize better to translations from language B when the two are typologically similar. Figure~\ref{fig:cross-eval-heatmaps} shows cross-evaluation perplexity matrices; Figure~\ref{fig:cross-eval-correlations} reports the correlation between syntactic similarity and cross-evaluation perplexity. 

Across both training dataset scales, we find a consistent and statistically significant negative correlation between typological similarity and cross-evaluation perplexity: the more syntactically similar two source languages are, the lower the perplexity when a model trained on the translation from one is evaluated on the translation from other. For the 100MB models (322 source language pairs), the correlation is moderate and significant (Pearson $r=-0.251$, $p<0.001$; Spearman $\rho=-0.244$, $p<0.001$),\footnote{We note that the language pairs are not statistically independent, and p-values should be interpreted cautiously here.} and a similar pattern holds for the 1GB models (226 pairs; Pearson $r=-0.228$, $p<0.001$; Spearman $\rho=-0.188$, $p<0.01$). The effect appears slightly stronger at 100MB, suggesting that smaller models are more sensitive to typological proximity. At 1GB, the effect remains robust but somewhat weaker. 

As expected, the diagonal of the heatmaps in Figure~\ref{fig:cross-eval-heatmaps} shows the lowest perplexities, with models performing best on the source language corpus they were trained on. Beyond the diagonal, related languages tend to show relatively low mutual perplexity, showing that typological similarity matters not only from the source language to English, but also between translations from different source languages, indicating that similar languages result in similar translations into English, i.e., in similar dialects of \textit{translationese}.

The two lowest-resource languages stand out in the 100MB setting. Models trained on Upper Sorbian and Northern Sámi have high perplexities on nearly all evaluation languages, reflecting that their corpus is likely less diverse and possibly generally of lower quality. Interestingly, Upper Sorbian as an \emph{evaluation} target proves relatively easy to model across many training languages, a pattern that however does not hold for Northern Sámi.
Even the Telugu dataset stands out in Figure~\ref{fig:cross-eval-correlations} as it gets unusually high perplexities by most models. An inspection of this dataset suggests that this pattern is likely driven by the high density of named entities in the corpus translated from Telugu.\footnote{Using spaCy \citep{honnibal2020spacy} named entity recognition on the 10MB corpora, we find that Telugu has the highest number of \textit{PERSON} entities (45,654), followed by Kannada (41,700) and Urdu (40,819). These counts are substantially higher than in the English FineWeb subset (24,694).}
Models trained on translations from typologically and geographically closer languages such as Kannada, Tamil, and to a lesser extent Urdu, perform better on Telugu, possibly due to shared named entities and overlapping cultural vocabulary.

\section{Discussion}

We find that using translated corpora has an impoverishing effect, leading to lower language modeling scores on native English corpora and to lower grammatical performance in the (more realistic) setup with more training data. This highlights the trade-off inherent to using translated data: We get more data, which can be crucial in low-resource scenarios, but this data's language differs from native production and is less diverse. 

Our results suggest that cross-lingual generalization from translated data is shaped by both structural and corpus-level factors, as well as translation quality. In this section, we discuss implications of our findings and their relation to other studies. 

\paragraph{Choice of Source Language}
The choice of source language matters when building translated training corpora. How to best select it depends on the goal. For general language modeling (perplexity), typological similarity to English is not a reliable guide: some typologically distant languages lead to competitive results, suggesting that corpus characteristics such as domain coverage and lexical distribution also play an important role. That said, at sufficient data scale and excluding very small languages, similarity \textit{does} have a consistent effect; it is simply one factor among several. 

For grammatical competence, typological similarity is however important. For the 1GB training data regimes, syntactic similarity strongly and significantly predicts performance on BLiMP, particularly for agreement and long-distance dependencies. When translated corpora are used and it is a priotity to have strong and robust grammatical preferences, choosing a structurally similar source language for translated training data is beneficial. 

These findings align with prior work on cross-lingual transfer in other setups. \citet{rice-etal-2025-untangling} find that typology, lexical overlap and dataset properties together predict zero-shot POS tagging performance. \citet{lin-etal-2019-choosing} similarly show that phylogenetic similarity, typological features, lexical overlap, and data size all contribute to transfer in tagging and parsing. \citet{blaschke-etal-2025-analyzing} further show that syntactic similarity is most predictive for word-level tasks, while surface overlap matters more for topic classification. Work on bilingual pre-training \citep{malkin-etal-2022-balanced} shows that language structure, not just data size, shapes transfer, and that structurally central languages can act as strong donors. For translated pre-training data specifically, \citet{doshi-etal-2024-pretraining} also find that the source language matters and that translationese negatively affects linguistic acceptability judgments. 

\paragraph{Dialects of Translationese}
Our cross-translation perplexity scores also show that translated data creates a \say{dialect} of the target language, which depends on the source language. 
This supports \citet{gellerstam86}'s and \citet{koppel-ordan-2011-translationese}'s finding that the source language leaves a footprint in the translated text, 
which we show is measurable and survives both the translation process and LM training. This shows that similar source languages' structural properties shape generalization in language models in comparable ways.

The dialects shape model behavior, and understanding this effect can help guide data generation choices (e.g., model selection, prompting) and filtering for diversity. This finding indicates that combining several source languages could be an area worth exploring, as this has the potential to increase diversity in the language of the resulting corpus. 

\paragraph{Translation Performance}
The strongest predictor of language modeling performance is the source-to-English translation quality of the Gemma model used to construct FineTranslations, showing that there is potential that as machine translation advances, or even using a stronger translation model, language models trained on translated data could become significantly better. 
However, the translation quality does not correlate with BLiMP scores. For morphological and syntactic aspects, typological similarity remains the strongest predictor. 


\section{Conclusion}

This study examined how the source language of machine-translated training data affects English language modeling and grammatical preferences in small language models. 

Perplexity results showed that no translated corpus matched the English baseline and that source language choice clearly matters. Syntactic closeness to English was correlated with lower perplexity (at the larger data scale), but some distant languages performed as well as or better than closely related ones, showing that language modeling quality depends on more than syntactic similarity alone. Lexical diversity in the corpus was a stronger predictor in low-data settings, but the strongest correlation was observed with the translation quality scores of the model used to translate the corpus. 
For grammatical preferences, typological similarity to English was strongly correlated with accuracy (at the larger data scale), particularly for agreement phenomena and long-distance dependencies. Corpus size and translation quality scores did not have a significant impact. 
Cross-evaluation experiments showed that models trained on translations from one language assign lower perplexity to translations from a different language when the two languages are syntactically closer. Data translated from more similar languages thus results in similar dialects of \textit{translationese}, showing that the source language leaves a footprint that survives both the translation process and language model training. 

Taken together, our results show that the source language of translated training data shapes learned linguistic knowledge in systematic ways, and that for grammatical competence, typological proximity is central. Future work should further disentangle corpus effects from linguistic similarity, and examine whether these patterns hold for larger models and models trained on significantly more data. 

\section*{Limitations}
Our setup translates into English rather than into low-resource languages, which is the more common use case in multilingual NLP. Translations into a high-resource language like English tend to be more fluent and well-formed, since MT systems are generally stronger in this direction. Our results therefore represent something closer to an upper bound on translated data quality; effects may be more pronounced when translating into lower-resource target languages where MT quality is weaker and translationese artifacts are stronger. 

We only consider the FineTranslations corpus that was translated with one LLM. Other translation systems may introduce different patterns. 

We do not formally disentangle the effect of the source language's typological properties from the properties of the corpus produced by translating from it, which is relevant as the corpora in our study differ in many properties such as the lexical distribution, domain coverage, noise level, and available data volume. Isolating the contribution of typological structure would at least require a more carefully matched data collection.

An important limitation of translated corpora that we do not address in this study is that translated corpora inherit the content of their source documents, which does not reflect the cultural knowledge, named entities, or pragmatic conventions of the target language community. Native English corpora contain culturally grounded references, idioms, and discourse patterns that translated text tends to flatten or omit. Models trained on translated data may therefore lack not only linguistic naturalness but also the world knowledge that native speakers of the language commonly have. 

Lastly, our experiments use small language models trained on at most 1GB of text. It is not clear whether the patterns we observe generalize to the much larger models and datasets used in current practice. Larger models may be more robust to translationese effects, or may amplify them in ways our setup cannot detect. 

\section*{Ethical Considerations}

We release models trained on translated web data. Although the underlying data has been filtered by its creators, it may still contain content that is inappropriate, biased, or otherwise problematic. 
As a result, the models may reproduce such patterns. 
However, the models are intended solely for research purposes, and as they are small and trained on limited data, we consider the likelihood that anyone would deploy them in ways that cause significant real-world harm to be low. 

\section*{Acknowledgments}

I thank the anonymous reviewers as well as the LiU NLP group for their valuable feedback, and Jannis Vamvas for pointing out a mistake in an earlier version of this paper. 
This research was supported by TrustLLM funded by Horizon Europe GA 101135671. The computations were enabled by the National Academic Infrastructure for Supercomputing in Sweden (NAISS), partially funded by the Swedish Research Council through grant agreement no. 2022-06725. 

\bibliography{custom}

\appendix
\section{BLiMP Category Scores}
\label{sec:appendix_full_blimp}

We report the full accuracy scores for all BLiMP subcategories in Table~\ref{tab:blimp_by_category_1GB}. Table~\ref{tab:blimp_categories_combined} presents the correlation of subcategory scores with base corpus size, type-token ratio, and syntactic similarity measures.

We use the following abbreviations in Table~\ref{tab:blimp_by_category_1GB}. For detailed descriptions of each category, we refer the reader to \citet{warstadt-etal-2020-blimp-benchmark}. 
\begin{enumerate}[label=\textbullet, itemsep=0pt]
    \item AnaAgr – Anaphor Agreement
    \item IrrFrm – Irregular Forms
    \item DetAgr – Determiner–Noun Agreement
    \item SVAgr – Subject–Verb Agreement
    \item ArgStr – Argument Structure
    \item Ellips – Ellipsis
    \item Bindng – Binding
    \item CtlRai – Control and Raising
    \item Quantf – Quantifiers
    \item FilGap – Filler–Gap Dependencies
    \item NPILic – NPI Licensing
    \item IslEff – Island Effects
\end{enumerate}

\begin{table*}[t]
\centering
\begin{adjustbox}{width=\textwidth}
\begin{tabular}{lrrrrrrrrrrrrr}
\toprule
 & AnaAgr & IrrFrm  & DetAgr &  SVAgr &
ArgStr &  Ellips  & Bindng &  CtlRai &
Quantf &  FilGap  & NPILic  & IslEff &
Overall \\
\midrule
eng & 86.75 & 88.90 & 82.08 & 65.57 & 70.94 & 66.65 & 68.24 & 66.60 & 75.33 & 65.43 & 54.97 & 46.35 & 67.14 \\ \midrule
arb & 68.05 & 76.65 & 79.35 & 66.15 & 71.21 & 58.45 & 74.03 & 61.08 & 83.53 & 62.99 & 52.23 & 40.15 & 65.14 \\
ces & 86.90 & 89.75 & 76.90 & 72.62 & 72.07 & 65.40 & 72.73 & 63.70 & 63.90 & 65.11 & 44.60 & 49.08 & 66.08 \\
ekk & 76.70 & 86.70 & 75.05 & 68.03 & 71.91 & 66.35 & 68.13 & 61.72 & 66.60 & 58.51 & 50.57 & 44.50 & 63.98 \\
fao & 81.60 & 84.00 & 76.00 & 70.95 & 69.61 & 60.65 & 68.87 & 66.66 & 72.83 & 63.74 & 59.29 & 39.83 & 65.66 \\
fas & 67.75 & 74.55 & 78.83 & 64.78 & 70.83 & 58.60 & 69.01 & 60.66 & 67.95 & 59.77 & 50.79 & 44.49 & 63.38 \\
heb & 79.20 & 90.40 & 73.71 & 69.17 & 70.04 & 54.55 & 72.59 & 63.08 & 71.90 & 60.66 & 46.19 & 52.27 & 65.08 \\
hsb & 71.85 & 80.15 & 67.66 & 67.67 & 69.22 & 57.05 & 65.10 & 61.64 & 66.58 & 62.20 & 58.23 & 47.36 & 63.29 \\
hun & 84.55 & 90.35 & 77.58 & 70.27 & 71.58 & 58.90 & 68.53 & 63.78 & 60.02 & 60.90 & 50.17 & 49.01 & 65.11 \\
ind & 73.75 & 86.45 & 80.60 & 69.12 & 71.93 & 63.55 & 69.04 & 63.06 & 73.95 & 56.31 & 65.91 & 43.00 & 66.39 \\
isl & 84.05 & 86.05 & 79.65 & 71.08 & 70.90 & 71.75 & 70.64 & 65.08 & 76.33 & 63.31 & 45.01 & 41.48 & 65.68 \\
jav & 72.65 & 91.80 & 73.91 & 68.33 & 71.77 & 61.00 & 67.43 & 60.44 & 81.05 & 60.84 & 61.11 & 40.65 & 65.30 \\
kan & 73.60 & 87.50 & 80.97 & 72.05 & 73.58 & 65.75 & 73.27 & 63.40 & 66.42 & 60.87 & 59.27 & 43.16 & 66.83 \\
kin & 86.55 & 89.15 & 78.33 & 70.45 & 71.50 & 55.35 & 68.96 & 58.64 & 68.70 & 61.23 & 55.01 & 43.48 & 65.18 \\
kmr & 79.05 & 87.85 & 83.92 & 69.45 & 71.88 & 57.30 & 69.24 & 61.92 & 76.08 & 64.57 & 54.70 & 40.19 & 66.24 \\
mlt & 79.50 & 87.10 & 81.33 & 69.88 & 70.17 & 55.90 & 70.53 & 65.30 & 63.45 & 63.96 & 49.37 & 43.14 & 65.06 \\
sme & 75.50 & 83.85 & 80.81 & 67.68 & 69.38 & 65.45 & 65.69 & 62.40 & 89.05 & 64.99 & 65.79 & 42.04 & 67.26 \\
swe & 86.85 & 87.40 & 80.53 & 74.40 & 71.57 & 66.35 & 69.43 & 65.50 & 76.60 & 64.79 & 56.27 & 43.88 & 67.67 \\
swh & 84.35 & 77.05 & 80.64 & 68.12 & 72.92 & 59.80 & 70.69 & 59.72 & 80.90 & 61.81 & 54.37 & 42.41 & 66.00 \\
tam & 79.75 & 82.95 & 81.24 & 71.57 & 73.29 & 57.20 & 71.67 & 65.28 & 69.42 & 59.53 & 62.23 & 48.54 & 67.54 \\
tel & 77.45 & 80.00 & 81.46 & 71.68 & 71.60 & 62.10 & 71.63 & 67.10 & 71.42 & 62.00 & 58.20 & 44.42 & 66.94 \\
ukr & 83.00 & 89.95 & 72.28 & 69.73 & 72.09 & 70.65 & 68.86 & 61.38 & 71.92 & 63.24 & 48.87 & 41.21 & 64.53 \\
urd & 76.45 & 77.00 & 75.25 & 68.45 & 70.81 & 56.10 & 70.20 & 63.40 & 72.38 & 62.59 & 58.94 & 45.14 & 65.36 \\
zsm & 72.15 & 84.20 & 80.62 & 67.68 & 72.19 & 60.85 & 68.61 & 63.08 & 69.50 & 58.84 & 63.31 & 44.42 & 65.96 \\
zul & 81.35 & 93.10 & 79.44 & 72.63 & 71.27 & 57.40 & 71.47 & 61.16 & 81.00 & 60.60 & 59.57 & 42.36 & 66.96 \\
\bottomrule
\end{tabular}
\end{adjustbox}
\begin{adjustbox}{width=\textwidth}
\begin{tabular}{lrrrrrrrrrrrrr}
\toprule
 & AnaAgr & IrrFrm  & DetAgr &  SVAgr &
ArgStr &  Ellips  & Bindng &  CtlRai &
Quantf &  FilGap  & NPILic  & IslEff &
Overall \\
\midrule
eng & 97.60 & 84.30 & 91.22 & 84.80 & 78.34 & 80.10 & 76.53 & 75.08 & 77.80 & 72.64 & 62.31 & 52.50 & 75.44 \\ \midrule
arb & 89.45 & 89.50 & 88.12 & 81.10 & 77.27 & 72.45 & 77.03 & 68.82 & 64.17 & 66.26 & 54.79 & 46.52 & 70.89 \\
ces & 97.55 & 93.70 & 90.10 & 85.40 & 79.11 & 80.60 & 75.09 & 71.26 & 61.45 & 65.79 & 53.94 & 57.30 & 73.33 \\
ekk & 95.60 & 92.60 & 86.95 & 81.82 & 78.32 & 74.00 & 77.37 & 70.54 & 68.33 & 64.90 & 52.70 & 54.76 & 72.31 \\
fao & \textcolor{lightgray}{90.80} & \textcolor{lightgray}{87.25} & \textcolor{lightgray}{82.51} & \textcolor{lightgray}{81.13} & \textcolor{lightgray}{74.34} & \textcolor{lightgray}{74.60} & \textcolor{lightgray}{68.99} & \textcolor{lightgray}{69.12} & \textcolor{lightgray}{74.95} & \textcolor{lightgray}{66.70} & \textcolor{lightgray}{63.97} & \textcolor{lightgray}{45.85} & \textcolor{lightgray}{70.61} \\
fas & 84.20 & 84.25 & 88.75 & 81.58 & 77.03 & 73.75 & 77.56 & 68.28 & 59.65 & 62.64 & 58.86 & 52.98 & 71.26 \\
heb & 95.50 & 90.55 & 88.30 & 84.37 & 79.30 & 80.95 & 74.77 & 66.22 & 62.62 & 67.81 & 56.50 & 57.83 & 73.11 \\
hsb & \textcolor{lightgray}{73.20} & \textcolor{lightgray}{82.40} & \textcolor{lightgray}{65.88} & \textcolor{lightgray}{68.17} & \textcolor{lightgray}{68.46} & \textcolor{lightgray}{50.40} & \textcolor{lightgray}{66.47} & \textcolor{lightgray}{60.12} & \textcolor{lightgray}{68.15} & \textcolor{lightgray}{62.00} & \textcolor{lightgray}{53.99} & \textcolor{lightgray}{49.32} & \textcolor{lightgray}{62.82} \\
hun & 96.50 & 96.70 & 86.83 & 84.28 & 79.36 & 76.45 & 76.06 & 71.04 & 62.27 & 65.70 & 56.96 & 54.38 & 72.90 \\
ind & 94.55 & 91.25 & 86.75 & 83.92 & 78.10 & 78.80 & 72.73 & 69.48 & 61.83 & 63.36 & 69.19 & 49.35 & 72.48 \\
isl & 97.55 & 85.20 & 87.85 & 84.82 & 76.48 & 76.60 & 71.81 & 72.98 & 72.25 & 66.31 & 63.04 & 54.36 & 73.37 \\
jav & 85.15 & 86.65 & 83.41 & 80.20 & 77.21 & 73.90 & 69.33 & 66.18 & 61.55 & 60.70 & 65.17 & 42.70 & 68.95 \\
kan & 93.50 & 89.10 & 86.51 & 83.58 & 79.03 & 74.30 & 72.10 & 71.66 & 61.52 & 64.40 & 61.53 & 54.15 & 72.28 \\
kin & \textcolor{lightgray}{96.25} & \textcolor{lightgray}{91.05} & \textcolor{lightgray}{87.65} & \textcolor{lightgray}{82.27} & \textcolor{lightgray}{75.76} & \textcolor{lightgray}{67.25} & \textcolor{lightgray}{73.40} & \textcolor{lightgray}{63.50} & \textcolor{lightgray}{74.38} & \textcolor{lightgray}{64.69} & \textcolor{lightgray}{64.43} & \textcolor{lightgray}{47.69} & \textcolor{lightgray}{71.64} \\
kmr & 91.50 & 88.95 & 90.00 & 84.80 & 76.61 & 73.65 & 75.13 & 65.40 & 72.30 & 67.54 & 65.59 & 46.10 & 72.68 \\
mlt & 93.60 & 88.10 & 88.44 & 84.35 & 77.86 & 74.15 & 75.40 & 69.20 & 62.15 & 67.93 & 52.51 & 52.73 & 71.84 \\
sme & \textcolor{lightgray}{83.05} & \textcolor{lightgray}{88.30} & \textcolor{lightgray}{78.59} & \textcolor{lightgray}{69.82} & \textcolor{lightgray}{70.26} & \textcolor{lightgray}{66.20} & \textcolor{lightgray}{66.74} & \textcolor{lightgray}{60.34} & \textcolor{lightgray}{88.67} & \textcolor{lightgray}{61.96} & \textcolor{lightgray}{62.83} & \textcolor{lightgray}{41.16} & \textcolor{lightgray}{66.89} \\
swe & 96.80 & 92.30 & 89.91 & 86.95 & 77.87 & 82.95 & 73.84 & 74.26 & 68.97 & 70.16 & 60.33 & 52.45 & 74.37 \\
swh & 95.30 & 90.05 & 89.84 & 83.33 & 77.57 & 69.65 & 72.49 & 64.34 & 80.03 & 67.40 & 58.23 & 52.56 & 72.77 \\
tam & 92.35 & 93.45 & 89.03 & 82.87 & 79.56 & 74.20 & 75.43 & 69.52 & 62.02 & 66.29 & 64.47 & 54.92 & 73.49 \\
tel & 92.40 & 94.35 & 89.15 & 83.52 & 78.82 & 71.65 & 73.20 & 71.54 & 62.52 & 64.83 & 64.50 & 56.15 & 73.36 \\
ukr & 95.80 & 91.75 & 86.24 & 83.83 & 77.89 & 79.75 & 75.13 & 63.90 & 55.50 & 66.47 & 56.10 & 54.39 & 71.48 \\
urd & 95.55 & 90.20 & 88.34 & 83.70 & 77.30 & 75.31 & 72.75 & 68.06 & 71.85 & 65.97 & 56.26 & 58.51 & 73.14 \\
zsm & 95.50 & 89.65 & 88.00 & 81.62 & 78.48 & 74.50 & 74.44 & 67.58 & 63.67 & 64.46 & 66.33 & 50.51 & 72.43 \\
zul & \textcolor{lightgray}{92.45} & \textcolor{lightgray}{84.65} & \textcolor{lightgray}{86.44} & \textcolor{lightgray}{82.43} & \textcolor{lightgray}{76.28} & \textcolor{lightgray}{72.30} & \textcolor{lightgray}{75.37} & \textcolor{lightgray}{63.84} & \textcolor{lightgray}{64.20} & \textcolor{lightgray}{63.14} & \textcolor{lightgray}{64.03} & \textcolor{lightgray}{44.44} & \textcolor{lightgray}{70.46} \\
\bottomrule
\end{tabular}
\end{adjustbox}
\caption{BLIMP accuracy (\%) by linguistic category and language. Above: 100MB; Below: 1GB models.  Results in \textcolor{lightgray}{gray} indicate that the training corpora for these low-resource languages were smaller than 1GB.}
\label{tab:blimp_by_category_1GB}
\end{table*}

\begin{table*}[t]
\centering
\scriptsize
\begin{tabular}{lcccccc}
\toprule
 & \multicolumn{3}{c}{100MB ($n$=24)} & \multicolumn{3}{c}{1GB ($n$=19)} \\
\cmidrule(lr){2-4} \cmidrule(lr){5-7}
BLiMP Category & Corpus & TTR & lang2vec & Corpus & TTR & lang2vec \\
\midrule
Overall 
& $-0.09$ ($-0.09$) 
& $+0.11$ ($+0.08$) 
& $+0.06$ ($-0.10$) 
& $+0.01$ ($+0.08$) 
& $-0.32$ ($-0.13$) 
& $+0.61^{**}$ ($+0.32$) \\ \midrule
Anaphora Agr. & $-0.04$ ($-0.06$) & $+0.33$ ($+0.39$) & $+0.04$ ($+0.05$) & $-0.00$ ($+0.31$) & $-0.01$ ($+0.14$) & $+0.74^{***}$ ($+0.73$) \\
Irregular Forms & $-0.08$ ($+0.03$) & $-0.04$ ($+0.04$) & $+0.36$ ($+0.32$) & $+0.18$ ($+0.36$) & $+0.45$ ($+0.48$) & $+0.25$ ($+0.26$) \\
Det-Noun Agr. & $+0.03$ ($-0.10$) & $-0.04$ ($-0.05$) & $+0.01$ ($-0.15$) & $+0.02$ ($+0.00$) & $-0.46^{*}$ ($-0.32$) & $+0.34$ ($-0.13$) \\
Subj-Verb Agr. & $-0.13$ ($-0.18$) & $-0.09$ ($-0.09$) & $-0.23$ ($-0.14$) & $+0.16$ ($+0.13$) & $-0.24$ ($-0.15$) & $+0.74^{***}$ ($+0.71$) \\
Arg. Structure & $+0.13$ ($+0.32$) & $+0.50^{*}$ ($+0.68$) & $-0.26$ ($-0.24$) & $+0.05$ ($+0.22$) & $+0.31$ ($+0.33$) & $+0.13$ ($+0.21$) \\
Ellipsis & $+0.10$ ($+0.26$) & $+0.12$ ($+0.02$) & $+0.09$ ($-0.05$) & $+0.53^{*}$ ($+0.48$) & $+0.21$ ($+0.23$) & $+0.61^{**}$ ($+0.75$) \\
Binding & $-0.08$ ($-0.03$) & $+0.13$ ($+0.15$) & $+0.27$ ($+0.19$) & $+0.28$ ($+0.38$) & $-0.23$ ($-0.17$) & $+0.36$ ($+0.02$) \\
Control/Rais. & $-0.02$ ($+0.14$) & $+0.13$ ($+0.03$) & $+0.47^{*}$ ($+0.51$) & $+0.20$ ($+0.24$) & $-0.06$ ($-0.02$) & $+0.39$ ($+0.35$) \\
Quantifiers & $-0.21$ ($-0.20$) & $-0.22$ ($-0.27$) & $+0.33$ ($+0.45$) & $-0.37$ ($-0.29$) & $-0.45$ ($-0.39$) & $-0.04$ ($-0.04$) \\
Filler-Gap & $+0.23$ ($+0.27$) & $+0.41^{*}$ ($+0.47$) & $+0.41$ ($+0.35$) & $-0.09$ ($-0.12$) & $-0.39$ ($-0.35$) & $+0.66^{**}$ ($+0.49$) \\
NPI Licensing & $-0.16$ ($-0.35$) & $-0.05$ ($-0.05$) & $-0.36$ ($-0.37$) & $-0.03$ ($-0.18$) & $+0.22$ ($+0.18$) & $-0.35$ ($-0.38$) \\
Island Effects & $+0.13$ ($+0.34$) & $+0.04$ ($+0.04$) & $+0.20$ ($+0.12$) & $-0.07$ ($+0.15$) & $-0.35$ ($-0.18$) & $+0.53^{*}$ ($+0.26$) \\
\bottomrule
\end{tabular}
\caption{Correlations for all BLiMP categories across model sizes, including overall BLiMP accuracy (first row). Values are Pearson $r$ with Spearman $\rho$ in parentheses. Significance levels: $^{***}p<0.001$, $^{**}p<0.01$, $^{*}p<0.05$ (based on Pearson $p$-values). 100MB $n$=24; 1GB $n$=19 (excludes fao, hsb, kin, sme, zul).}
\label{tab:blimp_categories_combined}
\end{table*}

\section{Approximated Morphological Complexity in the Corpora}
\label{app:morph_complexity}

To approximate morphological complexity, we use the bigram-based measures proposed by \citet{poelman-etal-2025-confounding}, which aim to capture surface-form diversity (potentially) arising from morphological variation. 
\textit{Accessor Variety} \citep{b16a8dfc-cc40-3270-b226-ef816413eef1, 10.1162/089120104773633394} is defined as the number of distinct tokens that appear adjacent to a given token in the corpus. 
\textit{Entropic Efficiency} is the Shannon entropy of the token's contextual distribution, normalized by its accessor variety. It measures how uniformly the observed contexts of a token are distributed. 
The intuition behind these measures is that morphologically richer languages tend to produce more diverse and less predictable token sequences. 
\citet{poelman-etal-2025-confounding} show that such statistics can serve as proxies for morphological complexity without requiring explicit morphological annotation. 
It is however important to keep in mind that accessor variety and entropic efficiency capture surface-form diversity in token sequences rather than morphological structure directly. Both metrics are sensitive not only to morphological processes but also to lexical diversity, syntactic variability, and domain effects, which all are expected to vary substantially across the corpora in our study. High accessor variety, for example, may arise from productive morphology, but also from topical heterogeneity or larger vocabularies. Entropic efficiency similarly reflects how evenly contexts are distributed around tokens, which can be influenced by factors such as discourse structure and stylistic variation rather than morphological productivity alone.

We calculate accessor variety and entropic efficiency using the Goldfish tokenizer of each source language for the source corpora; for the translated corpora, we always use the same English Goldfish tokenizer that we used for our models. We use the repository by \citet{poelman-etal-2025-confounding} for calculating the measures. 
Results are given in Table~\ref{tab:av_entropy_languages}. 

\begin{table*}[h]
\centering
\begin{adjustbox}{width=0.99\textwidth}
\begin{tabular}{lcccc}
\toprule
 & \multicolumn{2}{c}{Source Corpus} & \multicolumn{2}{c}{Translated Corpus} \\
\cmidrule(lr){2-3} \cmidrule(lr){4-5}
 & Accessor Variety & Entropic Efficiency & Accessor Variety & Entropic Efficiency \\
\midrule
English  & 6.83 & 0.412 & -- & -- \\ \midrule
Arabic & 17.42 & 0.468 & 6.74 & 0.337 \\
Czech & 13.90 & 0.485 & 6.53 & 0.373 \\
Estonian & 11.84 & 0.548 & 6.68 & 0.362 \\
Faroese & 8.65 & 0.535 & 5.01 & 0.330 \\
Persian & 11.69 & 0.494 & 5.32 & 0.309 \\
Hebrew & 12.13 & 0.501 & 5.11 & 0.312 \\
Upper Sorbian & 8.48 & 0.438 & 5.74 & 0.330 \\
Hungarian & 17.47 & 0.527 & 6.40 & 0.369 \\
Indonesian & 8.89 & 0.347 & 6.59 & 0.342 \\
Icelandic & 11.18 & 0.536 & 5.63 & 0.355 \\
Javanese & 12.46 & 0.428 & 8.10 & 0.295 \\
Kannada & 13.83 & 0.575 & 7.03 & 0.323 \\
Kinyarwanda & 11.08 & 0.490 & 6.08 & 0.326 \\
Kurmanji Kurdish & 8.93 & 0.432 & 6.41 & 0.334 \\
Maltese & 14.22 & 0.456 & 5.34 & 0.350 \\
Northern Sami & 11.62 & 0.471 & 7.00 & 0.329 \\
Swedish & 9.21 & 0.513 & 5.84 & 0.375 \\
Swahili & 11.09 & 0.470 & 5.98 & 0.315 \\
Tamil & 15.22 & 0.540 & 7.65 & 0.314 \\
Telugu & 14.13 & 0.517 & 8.44 & 0.325 \\
Ukrainian & 15.80 & 0.449 & 6.15 & 0.349 \\
Urdu & 5.30 & 0.408 & 5.11 & 0.277 \\
Malay & 7.17 & 0.329 & 6.75 & 0.298 \\
Zulu & 18.57 & 0.534 & 6.13 & 0.353 \\
\bottomrule
\end{tabular}
\end{adjustbox}
\caption{Accessor variety and entropic efficiency, as suggested by \citet{poelman-etal-2025-confounding} as approximators of morphological complexity. Higher values indicate greater surface-form diversity.}
\label{tab:av_entropy_languages}
\end{table*}

\subsection{Source versus Translated Corpora}

Source and target corpus have a positive but consistently non-significant correlation. For accessor variety, Pearson $r=0.3324$ ($p=0.1125$) and Spearman $\rho = 0.3487$ ($p=0.0949$). For entropic efficiency, Pearson $r=0.3821$ ($0.0654$) and Spearman $\rho = 0.3000$ ($0.1544$).  

As discussed in Section~\ref{sec:rel_work}, translation tends to systematically reduce morphological variation, as rich inflectional systems are often mapped onto simpler target-language constructions. Table~\ref{tab:av_entropy_languages} confirms this: entropic efficiency is higher in the original English FineWeb than in all FineTranslations corpora, while accessor variety results are mixed.

\subsection{Correlation with Performance}

We report all correlation scores with BLiMP and perplexity scores in Table~\ref{tab:morpho_correlations}. 

\paragraph{BLiMP accuracy.}
Neither source-corpus measure predicts BLiMP accuracy at either model size. Source accessor variety and entropic efficiency are non-significant across 100MB and 1GB models alike, with inconsistent directions between Pearson and Spearman for entropic efficiency (100MB: $r = 0.137$, $p = 0.522$, $\rho = 0.188$, $p = 0.379$; 1GB: $r = 0.273$, $p = 0.258$, $\rho = 0.358$, $p = 0.132$).
For the translated corpus, accessor variety shows a significant positive relationship with BLiMP accuracy at 100MB ($r = 0.417$, $p = 0.043$, $\rho = 0.427$, $p = 0.037$), suggesting that more contextually diverse (and thus potentially more morphologically complex) translated data achieve slightly higher grammatical accuracy. This effect however does not replicate at 1GB ($r = -0.236$, $p = 0.330$, $\rho = -0.138$, $p = 0.574$), where the sign reverses, indicating it may be an artifact. The correlation of the entropic efficiency of the translated corpus with BLiMP scores is positive but non-significant at both scales. 

Examining the individual BLiMP categories (Table~\ref{tab:blimp_cat_morph_correlations}; translated corpus only), we observe that entropic efficiency generally shows more significant correlations than accessor variety. Accessor variety produces significant correlations for the 100MB models in two categories (Argument Structure and NPI Licensing) but none for the 1GB models.
In contrast, entropic efficiency produces several positive and significant correlations. For the 100MB models, significant results appear in three categories: Anaphora Agreement (significant only under Pearson correlation), Irregular Forms, and Binding. For the 1GB models, positive and significant correlations are also found in three categories: Anaphora Agreement, Ellipsis, and Control/Raising.
However, for NPI Licensing, the correlations are significantly \textit{negative} for both sizes.

\paragraph{FLORES+ and FineWeb perplexity.}
Source-corpus measures show no consistent relationship with perplexity for any domain or model size. Accessor variety and entropic efficiency from the source corpus are non-significant across all eight conditions (four domains $\times$ two model sizes).

For the translated corpus, accessor variety is generally non-significant, with a single exception: at 1GB on Wikibooks, $r = 0.534$ ($p = 0.018$), indicating that higher contextual diversity in the translated training data is associated with \emph{higher} (i.e.\ worse)  perplexity in this domain at larger scale. 

Entropic efficiency of the translated corpus is the most consistent predictor. At 100MB, there are significant negative correlations for Wikibooks ($\rho = -0.576$, $p = 0.003$), Wikivoyage ($r = -0.460$, $p = 0.024$; $\rho = -0.576$, $p = 0.003$), and FineWeb ($\rho = -0.542$, $p = 0.006$); Wikinews is the sole exception ($r = -0.153$, $p = 0.474$, $\rho = -0.289$, $p = 0.171$). The same pattern holds at 1GB: significant effects for Wikibooks ($r = -0.616$, $p = 0.005$; $\rho = -0.656$, $p = 0.002$), Wikivoyage ($r = -0.576$, $p = 0.010$; $\rho = -0.635$, $p = 0.003$), and FineWeb ($r = -0.506$, $p = 0.027$; $\rho = -0.526$, $p = 0.021$), with Wikinews again non-significant. 
 This means that across both scales, languages whose translated corpus has a more uniform contextual distribution tend to achieve lower perplexity on English text. 

\begin{table*}[htbp]
\centering
\small
\setlength{\tabcolsep}{4pt}
\begin{tabular}{lc rrrr}
\toprule
 & & \multicolumn{2}{c}{Source corpus} & \multicolumn{2}{c}{Target corpus} \\
\cmidrule(lr){3-4} \cmidrule(lr){5-6}
Evaluation & $n$ & AV & EE & AV & EE \\
\midrule
BLiMP (100MB) & 24 & \makecell[r]{$0.088$ / $-0.007$ \\ $(.682)$ / $(.974)$} & \makecell[r]{$0.137$ / $0.188$ \\ $(.522)$ / $(.379)$} & \makecell[r]{$0.417^{*}$ / $0.427^{*}$ \\ $(.043)$ / $(.037)$} & \makecell[r]{$0.091$ / $0.047$ \\ $(.672)$ / $(.826)$} \\
BLiMP (1GB) & 19 & \makecell[r]{$-0.241$ / $-0.191$ \\ $(.320)$ / $(.433)$} & \makecell[r]{$0.273$ / $0.358$ \\ $(.258)$ / $(.132)$} & \makecell[r]{$-0.236$ / $-0.138$ \\ $(.330)$ / $(.574)$} & \makecell[r]{$0.321$ / $0.260$ \\ $(.181)$ / $(.283)$} \\
\midrule
Wikibooks (100MB) & 24 & \makecell[r]{$-0.260$ / $-0.340$ \\ $(.220)$ / $(.104)$} & \makecell[r]{$-0.036$ / $-0.040$ \\ $(.866)$ / $(.853)$} & \makecell[r]{$0.014$ / $0.043$ \\ $(.948)$ / $(.842)$} & \makecell[r]{$-0.283$ / $-0.576^{**}$ \\ $(.180)$ / $(.003)$} \\
Wikivoyage (100MB) & 24 & \makecell[r]{$-0.231$ / $-0.253$ \\ $(.278)$ / $(.233)$} & \makecell[r]{$-0.067$ / $-0.062$ \\ $(.755)$ / $(.774)$} & \makecell[r]{$0.095$ / $0.188$ \\ $(.659)$ / $(.379)$} & \makecell[r]{$-0.460^{*}$ / $-0.576^{**}$ \\ $(.024)$ / $(.003)$} \\
Wikinews (100MB) & 24 & \makecell[r]{$-0.258$ / $-0.242$ \\ $(.223)$ / $(.255)$} & \makecell[r]{$-0.061$ / $0.017$ \\ $(.778)$ / $(.939)$} & \makecell[r]{$0.002$ / $0.046$ \\ $(.994)$ / $(.832)$} & \makecell[r]{$-0.153$ / $-0.289$ \\ $(.474)$ / $(.171)$} \\
FineWeb (100MB) & 24 & \makecell[r]{$-0.236$ / $-0.226$ \\ $(.267)$ / $(.288)$} & \makecell[r]{$-0.068$ / $-0.059$ \\ $(.752)$ / $(.785)$} & \makecell[r]{$-0.040$ / $-0.029$ \\ $(.853)$ / $(.893)$} & \makecell[r]{$-0.286$ / $-0.542^{**}$ \\ $(.175)$ / $(.006)$} \\
\midrule
Wikibooks (1GB) & 19 & \makecell[r]{$0.058$ / $0.042$ \\ $(.814)$ / $(.864)$} & \makecell[r]{$0.013$ / $-0.053$ \\ $(.958)$ / $(.831)$} & \makecell[r]{$0.534^{*}$ / $0.431$ \\ $(.018)$ / $(.066)$} & \makecell[r]{$-0.616^{**}$ / $-0.656^{**}$ \\ $(.005)$ / $(.002)$} \\
Wikivoyage (1GB) & 19 & \makecell[r]{$0.108$ / $0.098$ \\ $(.660)$ / $(.689)$} & \makecell[r]{$0.073$ / $0.030$ \\ $(.767)$ / $(.904)$} & \makecell[r]{$0.376$ / $0.357$ \\ $(.112)$ / $(.133)$} & \makecell[r]{$-0.576^{**}$ / $-0.635^{**}$ \\ $(.010)$ / $(.003)$} \\
Wikinews (1GB) & 19 & \makecell[r]{$0.119$ / $0.124$ \\ $(.629)$ / $(.614)$} & \makecell[r]{$0.151$ / $0.198$ \\ $(.537)$ / $(.418)$} & \makecell[r]{$0.379$ / $0.308$ \\ $(.110)$ / $(.200)$} & \makecell[r]{$-0.106$ / $-0.101$ \\ $(.667)$ / $(.681)$} \\
FineWeb (1GB) & 19 & \makecell[r]{$0.173$ / $0.253$ \\ $(.480)$ / $(.297)$} & \makecell[r]{$0.121$ / $0.042$ \\ $(.623)$ / $(.864)$} & \makecell[r]{$0.318$ / $0.240$ \\ $(.185)$ / $(.321)$} & \makecell[r]{$-0.506^{*}$ / $-0.526^{*}$ \\ $(.027)$ / $(.021)$} \\
\bottomrule
\end{tabular}
\caption{Pearson $r$ and Spearman $\rho$ correlations between Accessor Variety (AV) and Entropic Efficiency (EE) from source and target corpora and model evaluation metrics. Higher BLiMP accuracy and lower FLORES/FineWeb perplexity indicate better performance. Superscripts: $^{*}p<.05$, $^{**}p<.01$.}
\label{tab:morpho_correlations}
\end{table*}

\begin{table*}[htbp]
\centering
\small
\setlength{\tabcolsep}{4pt}
\begin{tabular}{l rr rr}
\toprule
 & \multicolumn{2}{c}{100MB ($n=24$)} & \multicolumn{2}{c}{1GB ($n=19$)} \\
\cmidrule(lr){2-3} \cmidrule(lr){4-5}
Category & AV & EE & AV & EE \\
\midrule
AnaAgr & \makecell[r]{$0.108$ / $0.067$ \\ (.617) / (.754)} & \makecell[r]{$0.448^{*}$ / $0.349$ \\ (.028) / (.094)} & \makecell[r]{$-0.063$ / $-0.138$ \\ (.799) / (.574)} & \makecell[r]{$0.671^{**}$ / $0.691^{**}$ \\ (.002) / (.001)} \\
IrrFrm & \makecell[r]{$-0.210$ / $-0.264$ \\ (.324) / (.213)} & \makecell[r]{$0.519^{**}$ / $0.533^{**}$ \\ (.009) / (.007)} & \makecell[r]{$0.144$ / $0.146$ \\ (.555) / (.552)} & \makecell[r]{$0.424$ / $0.411$ \\ (.070) / (.081)} \\
DetAgr & \makecell[r]{$0.264$ / $0.376$ \\ (.213) / (.070)} & \makecell[r]{$0.080$ / $-0.019$ \\ (.710) / (.929)} & \makecell[r]{$-0.386$ / $-0.258$ \\ (.102) / (.286)} & \makecell[r]{$-0.080$ / $-0.165$ \\ (.746) / (.500)} \\
SVAgr & \makecell[r]{$0.165$ / $0.087$ \\ (.441) / (.686)} & \makecell[r]{$-0.190$ / $-0.110$ \\ (.374) / (.607)} & \makecell[r]{$-0.236$ / $-0.341$ \\ (.331) / (.153)} & \makecell[r]{$0.406$ / $0.411$ \\ (.085) / (.081)} \\
ArgStr & \makecell[r]{$0.505^{*}$ / $0.598^{**}$ \\ (.012) / (.002)} & \makecell[r]{$0.003$ / $-0.014$ \\ (.987) / (.949)} & \makecell[r]{$0.194$ / $0.211$ \\ (.426) / (.385)} & \makecell[r]{$0.334$ / $0.356$ \\ (.163) / (.135)} \\
Ellips & \makecell[r]{$0.048$ / $0.029$ \\ (.822) / (.892)} & \makecell[r]{$0.058$ / $0.016$ \\ (.788) / (.941)} & \makecell[r]{$-0.152$ / $-0.050$ \\ (.534) / (.839)} & \makecell[r]{$0.479^{*}$ / $0.461^{*}$ \\ (.038) / (.047)} \\
Bindng & \makecell[r]{$0.109$ / $0.029$ \\ (.612) / (.892)} & \makecell[r]{$0.491^{*}$ / $0.421^{*}$ \\ (.015) / (.040)} & \makecell[r]{$-0.352$ / $-0.291$ \\ (.140) / (.226)} & \makecell[r]{$0.205$ / $0.123$ \\ (.400) / (.616)} \\
CtlRai & \makecell[r]{$0.035$ / $-0.051$ \\ (.870) / (.813)} & \makecell[r]{$0.211$ / $0.252$ \\ (.321) / (.234)} & \makecell[r]{$0.106$ / $0.211$ \\ (.667) / (.385)} & \makecell[r]{$0.567^{*}$ / $0.604^{**}$ \\ (.011) / (.006)} \\
Quantf & \makecell[r]{$-0.216$ / $-0.238$ \\ (.312) / (.262)} & \makecell[r]{$0.274$ / $0.376$ \\ (.195) / (.070)} & \makecell[r]{$-0.158$ / $-0.160$ \\ (.518) / (.514)} & \makecell[r]{$0.140$ / $0.170$ \\ (.568) / (.486)} \\
FilGap & \makecell[r]{$0.202$ / $0.358$ \\ (.344) / (.086)} & \makecell[r]{$0.451^{*}$ / $0.395$ \\ (.027) / (.056)} & \makecell[r]{$-0.249$ / $-0.267$ \\ (.303) / (.270)} & \makecell[r]{$0.339$ / $0.281$ \\ (.155) / (.244)} \\
NPILic & \makecell[r]{$0.409^{*}$ / $0.405^{*}$ \\ (.047) / (.050)} & \makecell[r]{$-0.440^{*}$ / $-0.427^{*}$ \\ (.031) / (.037)} & \makecell[r]{$0.244$ / $0.252$ \\ (.315) / (.298)} & \makecell[r]{$-0.457^{*}$ / $-0.460^{*}$ \\ (.049) / (.048)} \\
IslEff & \makecell[r]{$-0.084$ / $-0.073$ \\ (.696) / (.734)} & \makecell[r]{$0.022$ / $-0.070$ \\ (.917) / (.745)} & \makecell[r]{$-0.298$ / $-0.091$ \\ (.216) / (.710)} & \makecell[r]{$0.291$ / $0.230$ \\ (.226) / (.344)} \\ \midrule
Overall & \makecell[r]{$0.417^{*}$ / $0.427^{*}$ \\ (.043) / (.037)} & \makecell[r]{$0.091$ / $0.047$ \\ (.672) / (.826)} & \makecell[r]{$-0.236$ / $-0.138$ \\ (.330) / (.574)} & \makecell[r]{$0.321$ / $0.260$ \\ (.181) / (.283)} \\
\bottomrule
\end{tabular}
\caption{Pearson $r$ and Spearman $\rho$ correlations between target-corpus Accessor Variety (AV) and Entropic Efficiency (EE) and BLiMP accuracy by category. Superscripts: $^{*}p<.05$, $^{**}p<.01$.}
\label{tab:blimp_cat_morph_correlations}
\end{table*}

\section{Gemma's Translation Quality and Perplexity in Source Languages}
\label{app:gemma_quality}

We report \textit{Gemma-3-27B}'s perplexity in each source language, along with chrF2 and BLEU scores for its translation quality into English, in Table~\ref{tab:translation_quality}. All scores are computed on the BOUQuET dataset \citep{theomnilingualmtteam2025bouquetdatasetbenchmarkopen}.
For the translation we use the following prompt: \textit{``Translate the following \{source language name\} text into English. Output only the English translation and nothing else.''}

\begin{table*}[t]
\centering
\begin{adjustbox}{width=0.8\columnwidth}
\begin{tabular}{lccc}
\toprule
Language & PPL & BLEU & chrF2 \\
\midrule
ces & 27.17 & 48.61 & 67.41 \\
ekk & 34.94 & 49.20 & 67.91 \\
fao & 95.16 & 33.01 & 52.75 \\
fas & 42.89 & 37.06 & 59.18 \\
heb & 27.81 & 46.91 & 67.25 \\
hun & 25.10 & 47.18 & 66.30 \\
ind & 88.25 & 40.28 & 60.51 \\
isl & 27.72 & 45.16 & 63.52 \\
jav & 114.80 & 39.26 & 61.51 \\
kan & 75.31 & 24.25 & 47.18 \\
kin & 69.66 & 22.89 & 45.84 \\
kmr & 71.35 & 36.37 & 57.34 \\
mlt & 26.40 & 54.54 & 71.45 \\
sme & 338.22 & 9.15 & 29.02 \\
swe & 36.57 & 55.98 & 72.22 \\
swh & 40.58 & 44.34 & 64.76 \\
tam & 68.21 & 24.16 & 46.53 \\
tel & 36.74 & 27.46 & 50.16 \\
ukr & 25.78 & 45.45 & 63.70 \\
urd & 62.54 & 46.32 & 64.72 \\
zsm & 130.89 & 41.53 & 62.26 \\
zul & 46.95 & 38.95 & 58.86 \\
\bottomrule
\end{tabular}
\end{adjustbox}
\caption{BOUQuET perplexity (PPL) and translation quality into English (BLEU, chrF2) of Gemma-3-27B for the language included in BOUQuET. }
\label{tab:translation_quality}
\end{table*}

\section{Ablation: Disentangling Language Set from Model Size}
\label{sec:filtering_ablation}

The 1\,GB models are trained on a subset of 19 languages, excluding five languages (Faroese, Upper Sorbian, Kinyarwanda, Northern Sámi, and Zulu) whose base corpora do not reach 1\,GB. 
Any difference between the 100\,MB and 1\,GB correlation patterns could therefore reflect either the effect of increased training data, or the removal of these five low-resource languages, or both.
To disentangle these factors, we conduct an ablation in which we exclude the same five languages from the 100\,MB models, which gives us a \emph{filtered} 100\,MB condition that is evaluated on the identical language set as the 1\,GB models.

\paragraph{MT--perplexity correlations are robust to filtering.}
In Table~\ref{tab:ppl_correlations_withfiltered}, we see that 
the strong negative correlations between translation quality (BLEU, chrF2) and perplexity are largely preserved after filtering.
For Wikibooks, Wikivoyage, FLORES Average, and English FineWeb, both BLEU and chrF2 remain highly significant ($p < .01$ or better) in the filtered 100\,MB models.
Wikivoyage even strengthens slightly after filtering (e.g., BLEU Pearson $r$: $-.77$ $\to$ $-.82$; chrF2: $-.74$ $\to$ $-.83$), suggesting that the excluded languages introduced noise in this domain.
The notable exception is Wikinews, where filtering causes a substantial drop: BLEU Pearson $r$ falls from $-.70$ ($p < .001$) to $-.47$ ($p = .052$), and chrF2 from $-.74$ ($p < .001$) to $-.44$ ($p = .069$).
Wikinews thus appears to be the evaluation domain most sensitive to the low-resource outliers.
 
\paragraph{Corpus size effects were largely driven by the excluded languages.}
Table~\ref{tab:ppl_correlations_withfiltered} also shows that the modest corpus size--perplexity correlations observed in the full 100\,MB set weaken further after filtering.
The two significant Spearman correlations (Wikinews $\rho = -.45$, $p < .05$; FineWeb $\rho = -.47$, $p < .05$) both become non-significant ($\rho = -.23$ and $-.27$, respectively), and all marginal Pearson trends disappear.
This indicates that much of the corpus size signal was carried by the five excluded languages, which tend to have both smaller corpora and higher perplexity. 
 
\paragraph{TTR--perplexity: filtering weakens but does not eliminate the effect.}
The type--token ratio (TTR) correlations show a more mixed pattern.
The Pearson correlations remain significant after filtering for Wikibooks, Wikivoyage, FLORES Average, and English FineWeb (all $p < .05$), though with slightly reduced effect sizes.
However, the Spearman correlations (which were significant across all non-Wikinews domains in the full set) uniformly lose significance after filtering, suggesting that the rank-order relationship was more sensitive to the outlier languages.
Wikinews shows the strongest shift: the already weak trend ($r = -.38$, $\rho = -.25$) collapses ($r = -.17$, $\rho = +.02$). 
 
However, the filtered 100\,MB models still show significant TTR--perplexity correlations (Pearson), while the 1\,GB models show near-zero effects (e.g., Wikibooks: $r = -.57$ vs.\ $r = +.07$). 
Since both conditions are evaluated on the same 19 languages, this confirms that the disappearance of TTR effects at 1\,GB reflects a consequence of more training data, and is not an artifact of the language set. 
 
\paragraph{Typological similarity: the excluded languages suppressed the signal.}
In a pattern opposite to most other predictors, the lang2vec correlations \emph{strengthen} after filtering.
Wikivoyage Spearman increases from $\rho = -.29$ (non-significant) to $\rho = -.53$ ($p < .05$), and FineWeb Spearman from $\rho = -.34$ to $\rho = -.50$ ($p < .05$).
The filtered 100\,MB and 1\,GB models show closely aligned lang2vec patterns, both with significant correlations for several domains.

\paragraph{BLiMP accuracy.}
The BLiMP results are in Table~\ref{tab:blimp_withfiltered}).
Corpus-level predictors (corpus size, TTR, lang2vec) show weak, non-significant correlations with the BLiMP overall accuracy in both the full and filtered 100\,MB conditions, and filtering changes nothing meaningful. 
The more notable finding concerns translation quality: in the full 100\,MB set, both BLEU ($r = -.37$, $p < .10$) and chrF2 ($r = -.40$, $p < .10$) show marginally significant negative correlations with BLiMP accuracy.
After filtering, these trends disappear (BLEU $r = -.19$; chrF2 $r = -.23$), confirming that the borderline MT--BLiMP relationship was driven by the five excluded languages. 

\begin{table*}[t]
\centering
\scriptsize
\begin{adjustbox}{width=\textwidth}
\begin{tabular}{llccccccccccc}
\toprule
\multirow{3}{*}{Model} & \multirow{3}{*}{Predictor} 
& \multicolumn{5}{c}{Pearson $r$} 
& \multicolumn{5}{c}{Spearman $\rho$} \\
\cmidrule(lr){3-7} \cmidrule(lr){8-12}
& & Books & Voyage & News & All & FineWeb 
& Books & Voyage & News & All & FineWeb \\
\midrule
100MB & Corpus Size 
& $-0.35$ & $-0.29$ & $-0.31$ & $-0.34$ & $-0.34$
& $-0.40$ & $-0.31$ & $-0.45^{*}$ & $-0.39$ & $-0.47^{*}$ \\
(n=24) & TTR 
& $-0.60^{**}$ & $-0.56^{**}$ & $-0.38$ & $-0.56^{**}$ & $-0.51^{*}$
& $-0.48^{*}$ & $-0.42^{*}$ & $-0.25$ & $-0.44^{*}$ & $-0.45^{*}$ \\
 & lang2vec 
& $-0.02$ & $-0.17$ & $-0.07$ & $-0.11$ & $-0.17$
& $-0.11$ & $-0.29$ & $-0.03$ & $-0.22$ & $-0.34$ \\
 & BLEU 
& $-0.79^{***}$ & $-0.77^{***}$ & $-0.69^{***}$ & $-0.83^{***}$ & $-0.79^{***}$
& $-0.71^{***}$ & $-0.79^{***}$ & $-0.58^{**}$ & $-0.78^{***}$ & $-0.70^{***}$ \\
 & chrF2 
& $-0.81^{***}$ & $-0.74^{***}$ & $-0.73^{***}$ & $-0.83^{***}$ & $-0.80^{***}$
& $-0.69^{***}$ & $-0.78^{***}$ & $-0.57^{**}$ & $-0.77^{***}$ & $-0.68^{***}$ \\
\midrule
100MB filt. & Corpus Size 
& $-0.28$ & $-0.25$ & $-0.21$ & $-0.27$ & $-0.31$
& $-0.22$ & $-0.22$ & $-0.23$ & $-0.22$ & $-0.27$ \\
(n=19) & TTR 
& $-0.57^{*}$ & $-0.54^{*}$ & $-0.17$ & $-0.52^{*}$ & $-0.50^{*}$
& $-0.30$ & $-0.32$ & $+0.02$ & $-0.25$ & $-0.23$ \\
 & lang2vec 
& $-0.15$ & $-0.28$ & $-0.32$ & $-0.28$ & $-0.37$
& $-0.26$ & $-0.53^{*}$ & $-0.09$ & $-0.42$ & $-0.50^{*}$ \\
 & BLEU 
& $-0.69^{**}$ & $-0.82^{***}$ & $-0.47$ & $-0.77^{***}$ & $-0.69^{**}$
& $-0.62^{**}$ & $-0.81^{***}$ & $-0.48^{*}$ & $-0.77^{***}$ & $-0.61^{**}$ \\
 & chrF2 
& $-0.71^{**}$ & $-0.83^{***}$ & $-0.44$ & $-0.77^{***}$ & $-0.69^{**}$
& $-0.61^{**}$ & $-0.81^{***}$ & $-0.47^{*}$ & $-0.76^{***}$ & $-0.62^{**}$ \\
\midrule
1GB & Corpus Size 
& $-0.44$ & $-0.17$ & $-0.21$ & $-0.28$ & $-0.29$
& $-0.42$ & $-0.15$ & $-0.21$ & $-0.27$ & $-0.27$ \\
(n=19) & TTR 
& $+0.07$ & $-0.07$ & $+0.33$ & $+0.06$ & $-0.01$
& $+0.04$ & $-0.09$ & $+0.22$ & $+0.02$ & $-0.01$ \\
 & lang2vec 
& $-0.45$ & $-0.35$ & $-0.32$ & $-0.42$ & $-0.43$
& $-0.53^{*}$ & $-0.50^{*}$ & $-0.14$ & $-0.52^{*}$ & $-0.45$ \\
 & BLEU 
& $-0.79^{***}$ & $-0.87^{***}$ & $-0.70^{***}$ & $-0.81^{***}$ & $-0.76^{***}$
& $-0.86^{***}$ & $-0.90^{***}$ & $-0.59^{**}$ & $-0.89^{***}$ & $-0.76^{***}$ \\
 & chrF2 
& $-0.84^{***}$ & $-0.90^{***}$ & $-0.76^{***}$ & $-0.86^{***}$ & $-0.82^{***}$
& $-0.87^{***}$ & $-0.90^{***}$ & $-0.60^{**}$ & $-0.89^{***}$ & $-0.77^{***}$ \\
\bottomrule
\end{tabular}
\end{adjustbox}
\caption{Results including a 100MB group that excludes the small corpora fao, hsb, kin, sme, zul that were excluded from the correlation analyses for the 1GB model. 
Correlations between predictor variables and perplexity (FLORES+ subsets and FineWeb) across model configurations. Significance levels: $^{***}p<0.001$, $^{**}p<0.01$, $^{*}p<0.05$. The 1GB and the filtered 100\,MB models exclude fao, hsb, kin, sme, zul (small corpora). lang2vec analyses have reduced sample sizes due to missing data (100MB n=20, 100mb filtered/1GB n=17)}. 
\label{tab:ppl_correlations_withfiltered}
\end{table*}

\begin{table*}[t]
\centering
\begin{adjustbox}{width=2\columnwidth}
\begin{tabular}{lccccc}
\toprule
 & Corpus Size & TTR & lang2vec & BLEU & chrF2 \\
\midrule
100MB (n=24) & $-0.09$ ($-0.10$) & $+0.11$ ($+0.08$) & $+0.06$ ($-0.10$) & $-0.37$ ($-0.37$) & $-0.40$ ($-0.39$) \\
100MB filt. (n=19) & $-0.05$ ($-0.06$) & $+0.13$ ($+0.12$) & $+0.10$ ($-0.06$) & $-0.19$ ($-0.24$) & $-0.23$ ($-0.23$) \\
1GB (n=19) & $+0.01$ ($+0.08$) & $-0.32$ ($-0.13$) & $+0.61^{**}$ ($+0.32$) & $+0.10$ ($+0.12$) & $+0.04$ ($+0.10$) \\
\bottomrule
\end{tabular}
\end{adjustbox}
\caption{Results including a 100MB group that excludes the small corpora fao, hsb, kin, sme, zul that were excluded from the correlation analyses for the 1GB model.  Correlations with overall BLiMP accuracy (Pearson $r$, Spearman $\rho$ in parentheses) across model configurations. The 1GB and filtered 100\,MB models exclude fao, hsb, kin, sme, zul. Significance levels: $^{***}p<0.001$, $^{**}p<0.01$, $^{*}p<0.05$. lang2vec analyses have reduced sample sizes due to missing data (100MB n=20, 100mb filtered/1GB n=17)}
\label{tab:blimp_withfiltered}
\end{table*}

\end{document}